\documentclass[10pt]{article} 
\usepackage[preprint]{tmlr}

\usepackage{amsmath,amsfonts,bm}

\def\eqref#1{equation~\ref{#1}}

\def\1{\bm{1}}

\DeclareMathAlphabet{\mathsfit}{\encodingdefault}{\sfdefault}{m}{sl}
\SetMathAlphabet{\mathsfit}{bold}{\encodingdefault}{\sfdefault}{bx}{n}

\usepackage{hyperref}
\usepackage{url}
\usepackage{graphicx}
\usepackage{tikz}
\usepackage{amsthm}
\usepackage{subcaption}
\usepackage{pifont}
\usepackage{multirow}
\usepackage{hhline}
\usepackage{paralist, tabularx}
\usepackage{soul}
\usepackage{tocloft}
\usepackage{color}
\usepackage{enumitem}
\definecolor{forestgreen}{rgb}{0.13, 0.55, 0.13}
\definecolor{bubblegum}{rgb}{0.99, 0.76, 0.8}
\definecolor{lightskyblue}{rgb}{0.53, 0.81, 0.98}
\definecolor{inchworm}{rgb}{0.7, 0.93, 0.36}
\usepackage{colortbl}
\definecolor{Red}{rgb}{139, 0, 0}
\definecolor{NavyBlue}{rgb}{0, 0, 128}
\definecolor{Green}{rgb}{0.13, 0.55, 0.13}

\definecolor{VividBlue}{HTML}{0055FF}      
\definecolor{VividCyan}{HTML}{00D4FF}      
\definecolor{DeepNavy}{HTML}{002266}      
\definecolor{VividOrange}{HTML}{FF6600}    
\definecolor{VividYellow}{HTML}{FFCC00}    
\definecolor{DeepRust}{HTML}{993300}       
\definecolor{VividGreen}{HTML}{487940}     
\definecolor{LightPurple}{HTML}{BD7AF7}      

\def\Monesymb{\textcolor{DeepNavy}{\ding{182}}\xspace}
\def\Mtwosymb{\textcolor{VividBlue}{\ding{183}}\xspace}
\def\Mthreesymb{\textcolor{VividCyan}{\ding{184}}\xspace}
\def\Mfoursymb{\textcolor{DeepRust}{\ding{185}}\xspace}
\def\Mfivesymb{\textcolor{VividOrange}{\ding{186}}\xspace}
\def\Msixsymb{\textcolor{VividYellow}{\ding{187}}\xspace}
\def\Msevensymb{\textcolor{teal}{\ding{188}}\xspace}
\def\Meightsymb{\textcolor{LightPurple}{\ding{189}}\xspace}

\def\Mone{{\hypersetup{hidelinks}\hyperref[subsubsec:withoutcot]{\Monesymb}}}
\def\Mtwo{{\hypersetup{hidelinks}\hyperref[subsubsec:textoutputcot]{\Mtwosymb}}}
\def\Mthree{{\hypersetup{hidelinks}\hyperref[subsubsec:inittraj]{\Mthreesymb}}}
\def\Mfour{{\hypersetup{hidelinks}\hyperref[subsec:langinteraction]{\Mfoursymb}}}
\def\Mfive{{\hypersetup{hidelinks}\hyperref[subsec:vlas]{\Mfivesymb}}}
\def\Msix{{\hypersetup{hidelinks}\hyperref[langactinteraction]{\Msixsymb}}}
\def\Mseven{{\hypersetup{hidelinks}\hyperref[subsec:onlytraining]{\Msevensymb}}}
\def\Meight{{\hypersetup{hidelinks}\hyperref[subsec:alsoinference]{\Meightsymb}}}

\newcommand{\Observation}[0]{\mathbf{X}}

\newcommand{\Prediction}[0]{\mathbf{O_{traj}}}
\newcommand{\InitialPrediction}[0]{\mathbf{O_{inittraj}}}
\newcommand{\PredictionRandvar}[0]{\mathrm{O_{traj}}}

\newcommand{\LatentRepresentation}[0]{\mathbf{Z}}
\newcommand{\textinput}[0]{\mathbf{T}}

\newcommand{\textoutput}[0]{\mathbf{O_{text}}}

\newcommand{\textoutputRandvar}[0]{\mathrm{O_{text}}}

\newcommand{\textoutputRandvari}[0]{\mathrm{O}^i_\mathrm{text}}
\newcommand{\textoutputj}[0]{\mathbf{O}^j_\mathbf{text}}
\newcommand{\textoutputi}[0]{\mathbf{O}^i_\mathbf{text}}

\newcommand{\textinputj}[0]{\mathbf{T}^j}

\newcommand{\FMObservation}[0]{\mathbf{X_{FM}}}

\expandafter 

\definecolor{LThree}{HTML}{A7D0E5}
\definecolor{LTwo}{HTML}{FFD9C6}
\definecolor{LOne}{HTML}{F4B4BC}
\definecolor{LNA}{HTML}{BFBFBF}
\definecolor{LUnknown}{HTML}{BFBFBF}

\newcommand{\CThreeTable}{\cellcolor{LThree}C3}
\newcommand{\CTwoTable}{\cellcolor{LTwo}C2}
\newcommand{\COneTable}{\cellcolor{LOne}C1}

\newcommand{\DThreeTable}{\cellcolor{LThree}D3}
\newcommand{\DTwoTable}{\cellcolor{LTwo}D2}
\newcommand{\DOneTable}{\cellcolor{LOne}D1}

\newcommand{\NATable}{\cellcolor{LNA}N/A}
\newcommand{\UnavailableTable}{\cellcolor{LNA}Unavailable}

\makeatletter
\newcommand{\DiagSplitCell}[5][1]{%
  \begin{tikzpicture}[baseline=(box.base)]
    \node[inner sep=0pt, outer sep=0pt, anchor=base,
          minimum width=\linewidth,
          minimum height=\dimexpr #1\ht\@arstrutbox + #1\dp\@arstrutbox\relax] (box) {};
    \path[fill=#2] (box.south west) -- (box.north west) -- (box.south east) -- cycle;
    \path[fill=#3] (box.north east) -- (box.north west) -- (box.south east) -- cycle;
    \node[anchor=south west, text=white, font=\bfseries, inner sep=1pt] at (box.south west) {#4};
    \node[anchor=north east, text=white, font=\bfseries, inner sep=1pt] at (box.north east) {#5};
  \end{tikzpicture}%
}
\makeatother

\newcommand{\cmark}{\ding{51}}%
\newcommand{\xmark}{\ding{55}}%
\newcommand{\blockcomment}[1]{}

\newcommand{\addition}[1]{\textcolor{blue}{#1}}
\newcommand{\removal}[1]{\textcolor{red}{\st{#1}}}

\usepackage{listings}

\newcommand{\KO}[1]{\textcolor{blue}{[Kemal: #1]}}

\usepackage{soul}

\usepackage{url}            
\usepackage{booktabs, arydshln}       
\usepackage{amsfonts}       
\usepackage{nicefrac}       
\usepackage{microtype}      

\makeatletter
\def\adl@drawiv#1#2#3{%
        \hskip.5\tabcolsep
        \xleaders#3{#2.5\@tempdimb #1{1}#2.5\@tempdimb}%
                #2\z@ plus1fil minus1fil\relax
        \hskip.5\tabcolsep}
\newcommand{\cdashlinelr}[1]{%
  \noalign{\vskip\aboverulesep
           \global\let\@dashdrawstore\adl@draw
           \global\let\adl@draw\adl@drawiv}
  \cdashline{#1}
  \noalign{\global\let\adl@draw\@dashdrawstore
           \vskip\belowrulesep}}
\makeatother

\usepackage[acronym,nowarn,section,nogroupskip,nonumberlist]{glossaries}

\glsdisablehyper{}
\newacronym{AD}{AD}{autonomous driving}
\newacronym{FM}{FM}{foundation model}
\newacronym{CoT}{CoT}{chain-of-thought}
\newacronym{LLM}{LLM}{large language model}
\newacronym{VLM}{VLM}{vision language model}
\newacronym{VLA}{VLA}{vision language action model}
\newacronym{BEV}{BEV}{bird's eye view}
\newacronym{VQA}{VQA}{visual question answering}
\newacronym{E2E}{E2E}{end-to-end}
\newacronym{MLP}{MLP}{multi-layer perceptron}
\newacronym{AR}{AR}{autoregressive}
\newacronym{LoRA}{LoRA}{low-rank adaptation}
\newacronym{RAG}{RAG}{retrieval-augmented generation}

\usepackage{titlecaps}
\usepackage{wrapfig}
\usetikzlibrary{mindmap, trees, shadows}

\newcommand*{\info}[5][16.3]{%
  \node (#5) [annotation, #3, scale=0.65, text width = #1em,
          inner sep = 2mm ] at (#2) {%
  \list{$\bullet$}{\topsep=0pt\itemsep=0pt\parsep=0pt
    \parskip=0pt\labelwidth=8pt\leftmargin=8pt
    \itemindent=0pt\labelsep=2pt}%
    #4
  \endlist
  };
}

\usepackage{blindtext}
\usepackage[capitalize]{cleveref}
\crefname{section}{Sec.}{Secs.}
\Crefname{section}{Section}{Sections}
\Crefname{table}{Table}{Tables}
\crefname{table}{Tab.}{Tabs.}

\title{Foundation Models for Trajectory Planning in Autonomous Driving: A Review of Progress and Open Challenges}

\author{\name Kemal Oksuz \email kemal.oksuz@bosch.com \\
     \addr Five AI Ltd., United Kingdom\\
     \addr Robert Bosch GmbH
     \AND
       \name Alexandru Buburuzan \email alexandru.buburuzan@bosch.com \\
     \addr Five AI Ltd., United Kingdom\\
     \addr Robert Bosch GmbH
     \AND
     \name Anthony Knittel \email anthony.knittel@bosch.com \\
     \addr Five AI Ltd., United Kingdom\\
     \addr Robert Bosch GmbH
     \AND
     \name Yuhan Yao \email yuhan.yao@de.bosch.com \\
     \addr Robert Bosch GmbH
     \AND
     \name Puneet K. Dokania \email dokania.puneet@bosch.com \\
      \addr Five AI Ltd., United Kingdom\\
     \addr Robert Bosch GmbH
      }

\usepackage[most]{tcolorbox}
\newtcolorbox{observation}[2][]{
  enhanced,
  title={\bfseries Summary of trade-offs: \textit{#2}},
  attach boxed title to top left={
    xshift=1.5em,
    yshift=-\tcboxedtitleheight/2
  },
  colback=gray!8,        
  colframe=gray!8,       
  boxrule=0.5pt,         
  coltitle=black,
  top=0.15in,
  boxed title style={
    size=small,
    colback=white,
    colframe=black,      
    boxrule=0.5pt        
  },
  drop shadow={white},   
  #1
}

\usepackage{etoolbox} 
\makeatletter
\newcommand\blfootnote[1]{%
  \begingroup
  \renewcommand\thefootnote{}%
  \@ifpackageloaded{hyperref}{\NoHyper}{}
    \footnote{#1}%
  \@ifpackageloaded{hyperref}{\endNoHyper}{}
  \addtocounter{footnote}{-1}
  \endgroup
}
\makeatother

\def\month{03}  
\def\year{2026} 
\def\openreview{\url{https://openreview.net/forum?id=E2L5J2O2Bk}} 

\begin{document}

\maketitle

\begin{abstract}
The emergence of multi-modal foundation models has markedly transformed the technology for autonomous driving, shifting away from conventional and mostly hand-crafted design choices towards unified, foundation-model-based approaches,  capable of directly inferring motion trajectories from raw sensory inputs. This new class of methods can also incorporate natural language as an additional modality, with Vision-Language-Action (VLA) models serving as a representative example. In this review, we provide a comprehensive examination of such methods through a unifying taxonomy to critically evaluate their architectural design choices, methodological strengths, and their inherent capabilities and limitations. Our survey covers 37 recently proposed approaches that span the landscape of trajectory planning with foundation models. Furthermore, we assess these approaches with respect to the openness of their source code and datasets, offering valuable information to practitioners and researchers. We provide an accompanying webpage that catalogues the methods based on our taxonomy, available at: \url{https://github.com/fiveai/FMs-for-driving-trajectories}.\blfootnote{The authors are from the ADAS Systems, Software \& Services Business Unit of Robert Bosch GmbH.}
\end{abstract}


\tableofcontents

\section{Introduction}\label{sec:introduction}
\Glspl{FM} are large-scale models that leverage vast amounts of data to learn representations that can be effectively adapted to a variety of downstream tasks. Depending on the type of data they process, these models are generally referred to by different terms. \Glspl{FM} that operate only on language data, such as BERT~\citep{bert}, GPT-2~\citep{gpt2}, Chat-GPT~\citep{chatgpt} and Qwen~\citep{qwen,qwen2,qwen25}, are known as \glspl{LLM}. Alternatively, models that process both language and visual data are considered \glspl{VLM}, with examples including CLIP~\citep{clip}, Flamingo~\citep{flamingo}, LLaVA~\citep{LLaVA,LLaVA15,LLaVAnext}, GPT-4o~\citep{gpt4}, Intern-VL~\citep{internvl, miniinternvl, internvl3} and Gemini~\citep{gemini}. As opposed to \glspl{LLM} and \glspl{VLM}, which generally output language data, a class of \glspl{FM} are designed to generate images from language inputs~\citep{imagen,diffusiontransformers}, or from both language and vision inputs~\citep{genie,genie2,genie3,Chameleon,cosmos}.

\begin{figure*}[t]
        \centering
        \includegraphics[width=\textwidth]{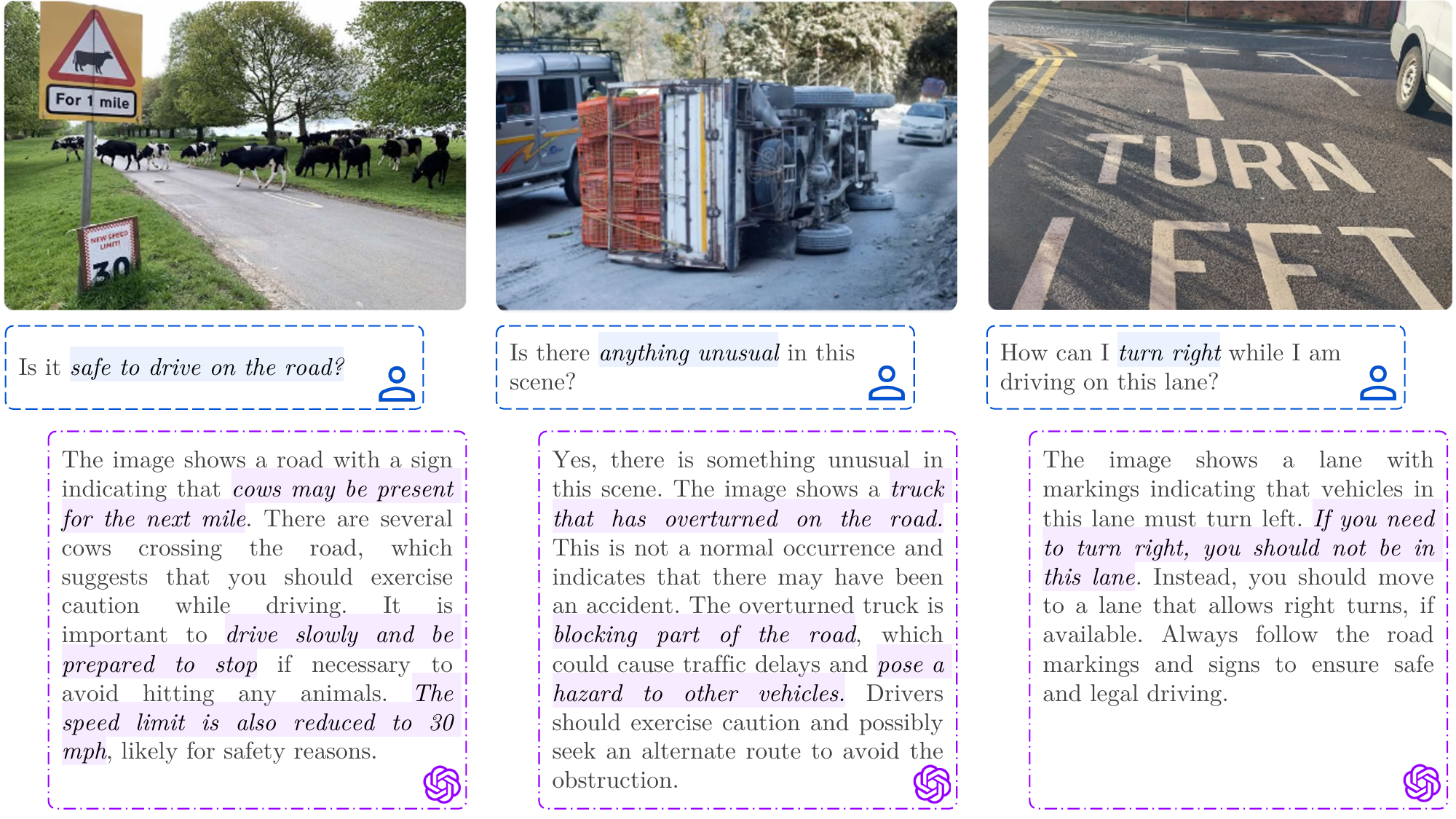}
        \caption{GPT-4o's response to driving-related prompts on three different scenarios.}
        \label{fig:motivation}
\end{figure*}

These models generally serve as a backbone upon which more specialised models for various different domains are built using fine-tuning; \gls{AD} is no exception. The scope of this work is to investigate the usefulness of \glspl{FM} for \gls{AD}. However, before delving into the \gls{AD}-specific approaches, a natural question to ask would be: ``\textit{Can off-the-shelf \glspl{FM}, specifically \glspl{VLM}, understand driving scenarios without being explicitly trained or fine-tuned for them?}''. To answer this, we prompt GPT-4o~\citep{gpt4} with three complex driving scenes and related questions, and show the interaction in \cref{fig:motivation}. The first two examples include rare driving scenes, where GPT-4o's replies are highly accurate and insightful. For example, in the first case, it not only understands the scene but also provides useful instructions on why and how one should drive cautiously in the given scenario. Similarly, in the last example, the model captures the misleading prompt and provides a reasonable explanation that promotes driving safely. Therefore, due to the scale of the architecture, training procedure, and the vast amount of training data involved, these models have learned representations that can already understand  driving scenes surprisingly well, making them a highly promising precursor to build \gls{AD}-specific solutions---a key driver to the paradigm shift currently observed in the \gls{AD} industry. Nevertheless, to deploy \glspl{FM} efficiently and reliably on edge devices for \gls{AD}, several aspects of customising such models (e.g., chain-of-thought inference cost, scale, accessibility to open weights, training/inference efficiency, obtaining suitable data for fine-tuning, etc.) form integral parts of the ongoing research and design choices in this area. Through this review, we present a holistic perspective on these aspects, highlighting progress, limitations, and open research directions.

\subsection{Scope and Contributions}
\label{subsec:scope}

\glspl{FM} can be used in several ways for \gls{AD}. Some methods leverage synthetic data for training~\citep{impromptuvla,trajectoryllm} and evaluation \citep{NeuroNCAP,DrivingSphere,pseudosimulation}, for which the models generating \gls{AD} data, such as \gls{AD} world models~\citep{gaia,drivedreamer,Vista,Panacea,DriveDreamer4D,DriveDreamer2}, can be helpful. A number of methods enhance scene understanding and reasoning capabilities of an \gls{FM} for \gls{AD} using \gls{VQA} tasks~\citep{humancentricautonomoussystemsllms,HolisticUnderstanding,dolphins,reason2drive,VLMdrivinglicense,agentthink,structuredlabeling}. Some methods also use an \gls{FM} to improve a specific capability of the \gls{AD} model such as perception~\citep{clipbevformer,ding2025hilm}, prediction~\citep{veon,autoocc} or control~\citep{copilot,languagempc}. Within this context, a group of methods leverage \glspl{FM} to yield textual actions, commonly by classifying the scene into one of the predefined meta-actions, such as \textit{``go straight''} and \textit{``slow down''}~\citep{chen2023endtoendembodieddecisionmaking,drivelikeahuman,safetyagent,dilu,alphadrive,lampilot,SurrealDriver,embodiedunderstanding,drivingeverywhere,RAD}. Last but not least, some models benefit from an \gls{FM} to directly operate the vehicle, typically through trajectory planning~\citep{VLP,VLMAD,drivevlm,ORION,omnidrive, emma,SimLingo}. 
While each of the above capabilities is essential for developing a robust and accurate \gls{AD} model, trajectory planning is, arguably, the most critical task for driving, where the others serve as auxiliary functions for this primary goal.
Consequently, considering the profound effect of \glspl{FM} on this crucial task, in this paper, we primarily focus on how trajectory planning models for \gls{AD} benefit from \glspl{FM}.\footnote{Although we use the term trajectory planning, our scope also includes the models predicting outputs in different forms such as control signals. One example of this is DriveGPT4~\citep{drivegpt4}, which predicts the target speed and turning angle of the autonomous vehicle.}
\begin{figure*}[t]
        \centering
        \includegraphics[width=0.92\textwidth]{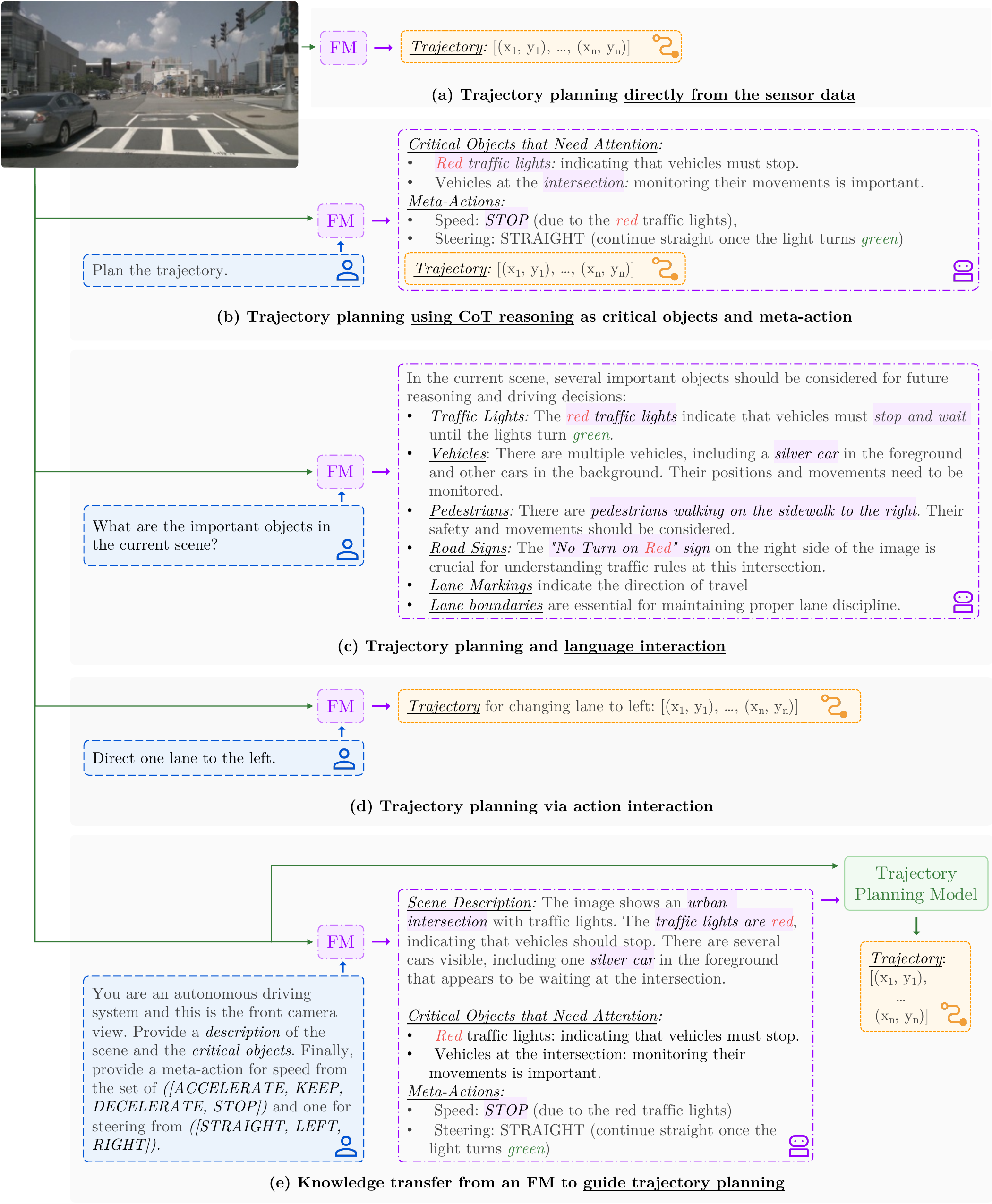}
        \caption{Different ways of how \glspl{FM} are helping trajectory planning. While \textbf{(a-d)} provide examples of \glspl{FM} tailored for trajectory planning, \textbf{(e)} is an example for an \gls{FM} guiding trajectory planning.
        The input image is from nuScenes~\citep{nuscenes}, the question is from DriveLM-nuScenes~\citep{drivelm}, the user instruction is from the SimLingo dataset~\citep{SimLingo}.}
        \label{fig:howfmshelp}
\end{figure*}

There are, in fact, various ways \glspl{FM} can be leveraged for enhanced trajectory planning, as illustrated in \cref{fig:howfmshelp}.
A common approach involves adapting existing \glspl{FM} through minimal architectural modifications followed by fine-tuning them on a task-specific dataset.
This fine-tuning can be as simple as training the model to output a trajectory directly from the sensor data~\citep{gptdriver,feedbackguided,ragdriver,DriveGPT4v2,S4Driver,autovla}(see \cref{fig:howfmshelp}(a)). 
Additionally, some models use \gls{CoT} reasoning, a technique that helps \glspl{LLM} break down a problem into multiple steps for enhanced reasoning.
\cref{fig:howfmshelp}(b) illustrates its common usage, in which the \gls{FM} first leverages \gls{CoT} in the form of critical objects and meta-actions, before producing the final trajectory.
The fact that \glspl{VLM} are equipped with a linguistic modality has also motivated several approaches that leverage this capability to enable language and/or action interactions with trajectory planning models~\citep{drivegpt4,drivelm,emma,lmdrive,SimLingo}.
As illustrated in \cref{fig:howfmshelp}(c), language interaction capability can offer reasoning behind a model's behaviour, giving reassurance to users. 
Differently, the action interaction capability facilitates driving assistance by executing driving commands of the user (refer to \cref{fig:howfmshelp}(d)), similar to existing \gls{VLA} models~\citep{rt2,openvla}. 
Alternatively, there are approaches where \glspl{FM} \textit{guide} an existing trajectory planning system by transferring their knowledge during training and/or inference~\citep{VLP,drivevlm,Senna,hedrive,vlme2e,diffvla,vdtauto,fasionad,VLMAD,dima,solve,DME-Driver}. 
\cref{fig:howfmshelp}(e) illustrates this set of approaches with an example, where a \gls{VLM} (GPT-4o~\citep{gpt4} in this case) outputs a description of the scene, critical objects that the ego should pay attention to, and a meta-action.

Although, as briefly indicated above, numerous studies benefit from \glspl{FM} for trajectory planning in autonomous driving, the field still lacks a holistic understanding of the progress achieved. The sheer variety of existing heterogeneous approaches has made it increasingly difficult to tell what truly distinguishes one method from another, since the underlying architectures, datasets, and training procedures vary widely and have great impact on the performance.
In this work, we aim to bring structure to this fragmented landscape by systematically analysing and comparing current techniques, highlighting the architectural design choices and capabilities that influence their effectiveness, and providing a unified perspective to guide further progress in applying foundation models to trajectory planning for autonomous driving. Our primary contributions, therefore, are:

\begin{enumerate}
    \item We introduce a hierarchical taxonomy of the methods that employ \glspl{FM} for trajectory planning in autonomous driving and systematically analyse 37 existing methods based on this taxonomy.
    \textit{Our study is a comprehensive effort to organise, interpret, and unify this rapidly evolving field}, where a variety of heterogenous approaches has emerged without clear benchmarking and understanding of their distinctions. We believe that our study provides a structured foundation to help the field in making methodical progress.
    \item In addition to \textit{providing practical guidance} on how to tailor and to fine-tune an FM for trajectory planning and strategies for curating datasets for different use cases, \textit{we also assess these approaches in terms of how open their code and data are}, in order to give practitioners and researchers useful pointers for their reproducibility and reuse. We also outline key future challenges and open research questions from multiple perspectives, including efficiency, robustness, evaluation benchmarks and sim-to-real transfer.
\end{enumerate}

Given that pioneering works in this domain first emerged as preprints in late 2023 (e.g., GPT-Driver~\citep{gptdriver}, DriveGPT4~\citep{drivegpt4}), this review encompasses approximately two years of research progress through October 2025 in leveraging \glspl{FM} for autonomous trajectory planning. Within this period, we select 37 methods published in peer-reviewed venues to provide a clear conceptual analysis and well-structured organisation. We also include a limited number of representative preprints to incorporate emergent research. For example, V2X-VLM~\citep{v2xvlm} represents a specialised \gls{FM} for trajectory planning uniquely utilising additional infrastructure images, while FASIONAD++\citep{fasionad} employs a \gls{FM} selectively to guide trajectory planning only when the \gls{AD} model has a high uncertainty. We emphasise that our objective is not to compare or rank these methods, but rather to organise and analyse them conceptually.

\subsection{Comparison with Previous Reviews}
\label{subsec:prev_reviews}
Considering that there are several reviews or surveys focusing on \gls{AD}, we would like to mention key differences between ours and the existing ones. 
One particular set of these reviews focuses on \gls{AD}, usually to present and discuss the state-of-the-art at the time they were published~\citep{IEEEADSurvey, selfdrivingcarsurvey,janai2021computervisionautonomousvehicles,Tampuu_2022,reviewe2eieeeaccess,li2023knowledgedrivenautonomousdriving,ADsystemsurvey, tpamie2esurvey}. Some of these surveys have a narrower scope such as focusing on \gls{E2E}-trainable models~\citep{Tampuu_2022,reviewe2eieeeaccess,tpamie2esurvey}, models using reinforcement learning~\citep{RLsurveyAD}, imitation learning-based approaches~\citep{ILsurveyAD}, or existing \gls{AD} datasets~\citep{liu2024surveyautonomousdrivingdatasets}. Alternatively, rather than trajectory planning, some reviews focus on a single auxiliary task for \gls{AD} such as perception~\citep{perceptionsurvey}, occupancy prediction~\citep{occpredsurvey} or motion planning~\citep{motionplanningsurvey}. While these are useful for their respective purposes, they do not pay special attention to how \glspl{FM} can be used within the context of trajectory planning. As a result, our review is complementary to this set of reviews.

Due to the common adoption of \glspl{FM} in several fields, including \gls{AD}, a few review papers consider how these models can help \gls{AD}~\citep{gao2024surveyfoundationmodelsautonomous, yang2024llm4drivesurveylargelanguage,yang2024llm4drivesurveylargelanguage,zhou2024visionlanguagemodelsautonomous,Review1LLMAD,CoTADsurvey}. Although they consider the use of \glspl{FM} for \gls{AD}, their scope is broader than ours. Specifically, these studies aim to cover how \glspl{FM} can be useful from various perspectives, including perception, data generation, scene understanding, as well as trajectory planning. Consequently, the discussion on trajectory planning is quite limited, whereas we focus exclusively on trajectory planning, allowing a more comprehensive discussion.
As one similar survey to ours, \cite{jiang2025surveyvisionlanguageactionmodelsautonomous} discuss different architectural paradigms for the models built on \glspl{VLM} and their chronological progress. In addition, we include knowledge transfer approaches using \glspl{FM} in our scope, introduce a hierarchical taxonomy over this broader set of models, delve deeper into the design choices for adapting a \gls{VLM} for trajectory planning and elaborate on the openness of the methods that can be useful for researchers and practitioners.

\section{Notations and Background}
\label{sec:background}
\label{subsec:notation}

\begin{figure*}[t]
        \centering
        \includegraphics[width=\textwidth]{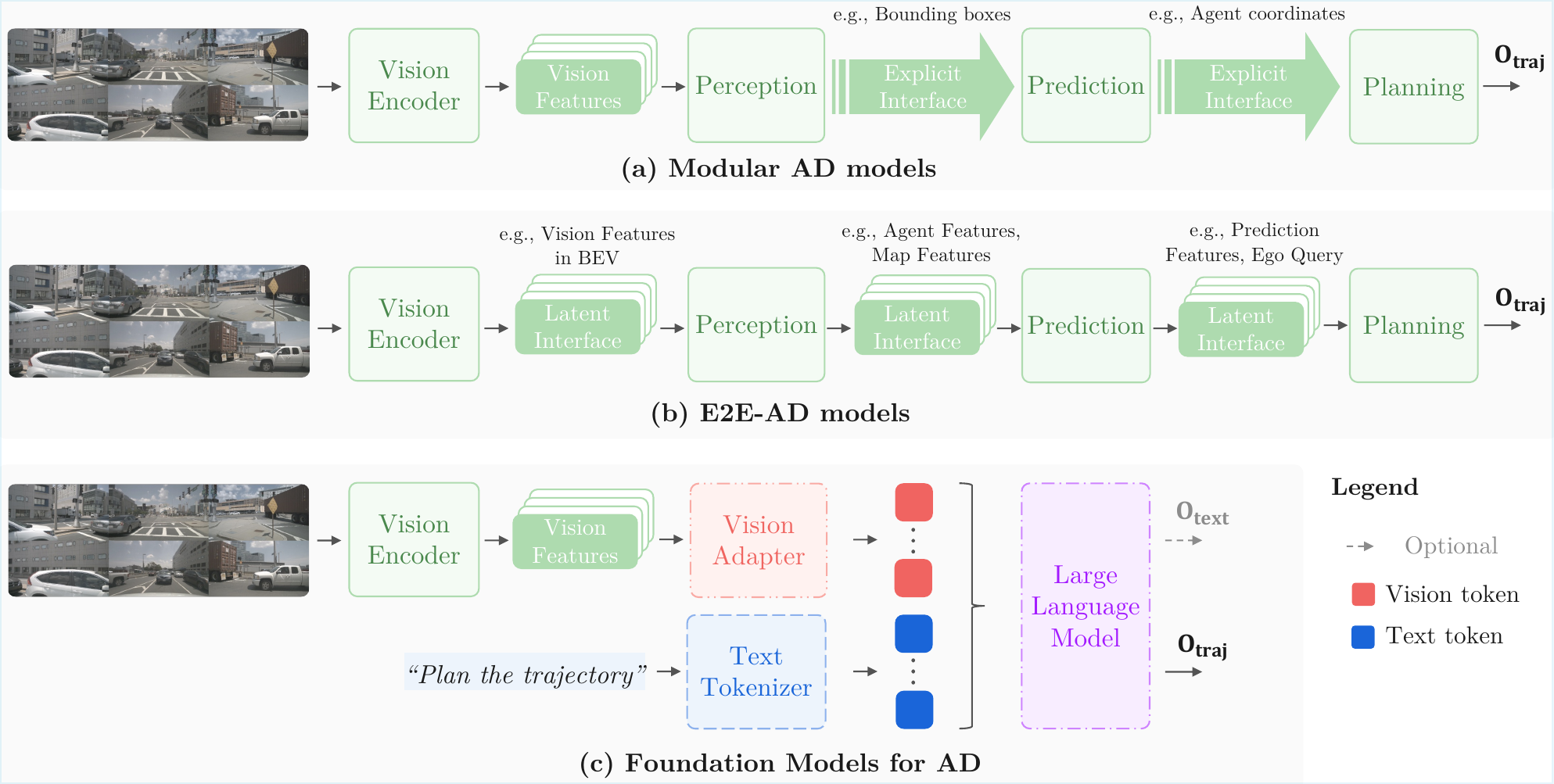}
        \caption{Trajectory planning methods. \textbf{(a)} \textbf{Modular} approaches use explicit interfaces between different modules. \textbf{(b)} \textbf{End-to-End} (\gls{E2E}-\gls{AD}) models replace explicit interfaces with latent ones, allowing all modules to be jointly differentiable. \textbf{(c)} \textbf{\gls{FM}-based} methods that follow the typical \gls{VLM} pipeline. The text output of the \gls{VLM} can also be used, e.g., \gls{CoT} reasoning. In this illustration, we simplify the pipelines to provide a high-level overview of how these models work, yet they can include different number of components and more complex connections across the modules. Input images are taken from nuScenes~\citep{nuscenes}.}
        \label{fig:adtypes}
\end{figure*}

\textbf{Notations.} We denote the set of observations by $\Observation$, where $\Observation$ typically includes a subset of (i) 
sensor readings generally from cameras, lidar, or radar, (ii) 
the state of the ego vehicle such as its velocity, acceleration and steering, and (iii) 
an indicator of the route, which can be in the form of a GPS target point or a high level command such as  \{go straight, turn left, turn right\}. 
We use $\Prediction \in \mathbb{R}^{k \times p}$ to represent the output trajectory, such that $k$ is the prediction horizon and $p$ is the dimension of the output at each time step, commonly $p=2$ for \gls{BEV} coordinates. 
Considering that the models typically aim to maximise the likelihood of the data, $\Prediction$ can be assumed to be a sample from the predictive distribution of the model, $\Prediction \sim p(\PredictionRandvar|\Observation)$, where $\PredictionRandvar$ denotes the corresponding random variable.
Furthermore, we use $\InitialPrediction$ to denote an initially-predicted trajectory to account for methods predicting an initial trajectory before refining it to obtain $\Prediction$.
$\textinput$ and $\textoutput$ denote the input text to the \gls{FM} and the output text from it, where each of them is a sequence of text tokens, i.e., $\textinput=\{\textinputj\}_{j=1}^n$ and $\textoutput=\{\textoutputj\}_{j=1}^m$ with $n$ and $m$ being their sequence lengths. 

\glspl{FM} typically generate the next token of a text output conditioning on the previous tokens, known as the \textit{\gls{AR} generation} or \textit{next token prediction}. In the context of \glspl{VLM}, which predict a categorical distribution over the text tokens, we formulate this as  $\textoutputi \sim p(\textoutputRandvari|\Observation,\textinput, \{\textoutputj\}_{j=1}^{i-1})$ with $\textoutputRandvari$ being the random variable for the $i$-th token. For clarity, when we refer to the entire text output, we use $\textoutput \sim p(\textoutputRandvar|\Observation,\textinput)$.
Additionally, in the case of knowledge transfer (refer to \cref{fig:howfmshelp}(e)), the set of observations provided to the \gls{FM} can be different from that provided to the trajectory planning model for \gls{AD}. For example, while the \gls{AD} model may receive multi-view images, the corresponding \gls{FM} may receive only the front image, potentially with additional privileged information, e.g., VLM-AD~\citep{VLMAD}. As a result, while discussing the approaches with knowledge transfer, we denote the observations provided to the \gls{FM} as $\FMObservation$, and corresponding latent representations as $\LatentRepresentation$.

\textbf{Trajectory planning for \gls{AD}.} 
The trajectory planning task for \gls{AD} is typically considered as a non-deterministic environment where the world is partially observed through the sensors. Reinforcement Learning~\citep{kazemkhani2024gpudrive} and Tree-Search~\citep{huang2024learning} methods typically formulate this environment as either a Fully- or Partially-Observed Markov Decision Process~\citep{russell1995modern}, and explicitly define numeric rewards and penalties to be optimised, while considering the distribution of expected rewards to form a control policy. In contrast, Imitation Learning delegates planning strategy to a teacher, where the objective of such a planner is to predict how a teacher model would act, and use this prediction as a plan for operating the vehicle. Essentially, given a set of observations $\Observation$ from the sensors, the trajectory planning task involves determining a trajectory for the ego vehicle to follow  ($\Prediction$). And, to operate a vehicle, $\Prediction$ is typically propagated to a controller such as a Proportional-Integral-Derivative (PID) controller~\citep{pidcontroller} to obtain control signals, such as for the accelerator and steering.

Model architectures for trajectory planning can be divided into three main groups, as shown in \cref{fig:adtypes}. Modular approaches, shown in \cref{fig:adtypes}(a), use explicit interfaces, where each module independently aims to solve a subproblem such as perception~\citep{DDETR,motr,petr, streampetr,saod,BRSLoss,bevformer}, prediction~\citep{mtr,mtr+,qcnet,forecastmae,emp} or planning~\citep{vectornet,drivingwithllms}. This architecture does not allow gradient flow during training, i.e., it is not \gls{E2E}-trainable, and each module is optimised for its own objective. \gls{E2E}-\gls{AD} models in \cref{fig:adtypes}(b) replace explicit interfaces by latent representations from the previous module, to optimise all modules jointly for trajectory planning~\citep{uniad,vad,vadv2,bevplanner,hydramdp,Paradrive,genad,DiffusionDrive,BridgeAD,MomAD}. 
\gls{FM}-based models are recent approaches that are built directly on \glspl{FM} to benefit from their vast world knowledge~\citep{drivegpt4,feedbackguided,omnidrive,SimLingo,emma,ORION,S4Driver,solve}. \cref{fig:adtypes}(c) illustrates a typical \gls{VLM} pipeline~\citep{qwenvl,blip2,LLaVA,LLaVAnext}, in which a vision encoder~\citep{ResNet,vit,swin} outputs vision features, which are then projected onto the \gls{LLM} embedding space using a vision adapter, e.g., a linear layer~\citep{LLaVA}. Finally, an \gls{LLM} processes vision and text tokens to yield the final trajectory. As more recent approaches, existing \gls{E2E}-\gls{AD} and \gls{FM}-based methods are predominantly trained using Imitation Learning.

\section{A Hierarchical Taxonomy} \label{sec:taxonomy}

\definecolor{green}{HTML}{AFD6AB}

\tikzset{
  basic/.style = {
    fill=green,
    rounded corners=3pt,
    font=\sffamily,
    align=center,
    rectangle
  },
  root/.style   = {basic, text width=5cm},
  level1/.style = {basic, text width=15em},
  level2/.style = {basic, text width=12.5em},
  level2_/.style = {basic, text width=12.5em},
}

\tikzset{
  every annotation/.style = {
    draw,
    fill=green,
    rounded corners=3pt,
    font=\sffamily,
    align=left
  }
}

\begin{figure*}[t!]
\hypersetup{hidelinks}
\centering
\begin{tikzpicture}[scale=0.85, transform shape,
  level 1/.style={sibling distance=9.5cm},
  level 2/.style={sibling distance=4.75cm},
  level 2_/.style={sibling distance=4.75cm},
  edge from parent/.style={->,draw},
  >=latex, every annotation/.style = {draw, fill = white, font = \large}]

\node[root] {FM-based methods for AD}
  child {node[level1] (builton) {\glspl{FM} tailored for trajectory planning}
      child {node[level2] (nointeraction) {Models focused solely on traj planning (\cref{subsec:trajplanningonly})}}
      child {node[level2_] (limitedinteraction) {Models providing additional capabilities (\cref{subsec:additionalcapabilities})}}
  }
  child {node[level1] (externalFM) {\glspl{FM} guiding trajectory planning} 
    child {node[level2] (onlytraining) {Knowledge distillation only during training (\cref{subsec:onlytraining})}}
    child {node[level2] (traintest) {Knowledge transfer\\during inference (\cref{subsec:alsoinference})}}
    };

\blockcomment{
\node[root] {\gls{AD} models using VLM/LLM}
  child {node[level1] (externalFM) {Models improving an existing AD stack} 
    child {node[level2] (onlytraining) {Knowledge distillation during training}}
    child {node[level2] (traintest) {Knowledge transfer during inference}}
    }
  child {node[level1] (builton) {Towards building VLAs: Models built on a VLM}
      child {node[level2] (nointeraction) {Models focusing on trajectory planning only}}
      child {node[level2] (limitedinteraction) {Models with language interaction (\gls{VQA})}}
      child {node[level2] (extendedinteraction) {Models with language \& action interaction}}
  };
}
 \info[19]{onlytraining.south}{anchor=north,xshift = 0em, yshift= 0em}{
      \item[\Mseven]
      \item[VLP \citep{VLP}]
      \item[VLM-AD \citep{VLMAD}]
      \item[DiMA \citep{dima}]
      \item[Solve-E2E \citep{solve}]
    }{onlytraining}

 \info[19]{traintest.south}{anchor=north,xshift = 0em, yshift= 0em}{
      \item[\Meight]
      \item[VLM-E2E \citep{vlme2e}]
      \item[DME-Driver \citep{DME-Driver}]
      \item[Senna-E2E \citep{Senna}]
      \item[DiffVLA \citep{diffvla}]
      \item[DriveVLM-Dual \citep{drivevlm}]
      \item[Solve-E2E-Async \citep{solve}]
      \item[DiMA-Dual \citep{dima}]
      \item[HE-Drive \citep{hedrive}]
      \item[VDT-Auto \citep{vdtauto}]
      \item[FASIONAD++ \citep{fasionad}]
      \item[AsyncDriver \citep{chen2024asynchronous}]
    }{traintest}

 \info[19]{nointeraction.south}{anchor=north,xshift = 0em, yshift= 0em}{%
      \item[\Mone\textbf{No \gls{CoT}}]
      \item[CarLLaVA \citep{CarLLaVA}]
      \item[DriveGPT4-v2 \citep{DriveGPT4v2}]
      \item[V2X-VLM \citep{v2xvlm}]
      \item[\Mtwo\textbf{Text as \gls{CoT}}]
      \item[GPT-Driver \citep{gptdriver}]
      \item[Drive-VLM \citep{drivevlm}]
      \item[Auto-VLA \citep{autovla}]
      \item[RAG-Driver \citep{ragdriver}]
      \item[S4-Driver \citep{S4Driver}]
      \item[\Mthree\textbf{Trajectory as \gls{CoT}}]
      \item[Agent-driver \citep{agentdriver}]
      \item[FeD \citep{feedbackguided}]
      \item[Solve-VLM \citep{solve}]
    }{nointeraction}

 \info[19]{limitedinteraction.south}{anchor=north,xshift = 0em, yshift= 0em}{    
    \item[\Mfour\textbf{Language interaction}]
    \item[DriveGPT4 \citep{drivegpt4}]
    \item[DriveLM-Agent \citep{drivelm}]
    \item[Emma \citep{emma}]
    \item[OpenDriveVLA \citep{OpenDriveVLA}] 
    \item[DiMA-MLLM \citep{dima}]
    \item[Omni-Q \citep{omnidrive}]
    \item[Omni-L \citep{omnidrive}]
    \item[Orion \citep{ORION}]
    \item[\Mfive\textbf{Action interaction}]
    \item[DriveMLM \citep{drivemlm}]
    \item[LMDrive \citep{lmdrive}]
    \item[\Msix\textbf{Language \& action interaction}]
    \item[SimLingo \citep{SimLingo}]
    }{limitedinteraction}
\end{tikzpicture}
\caption{Taxonomy of trajectory planning methods utilising or getting help from \glspl{FM}.
}
\label{fig:Taxonomy}
\end{figure*}
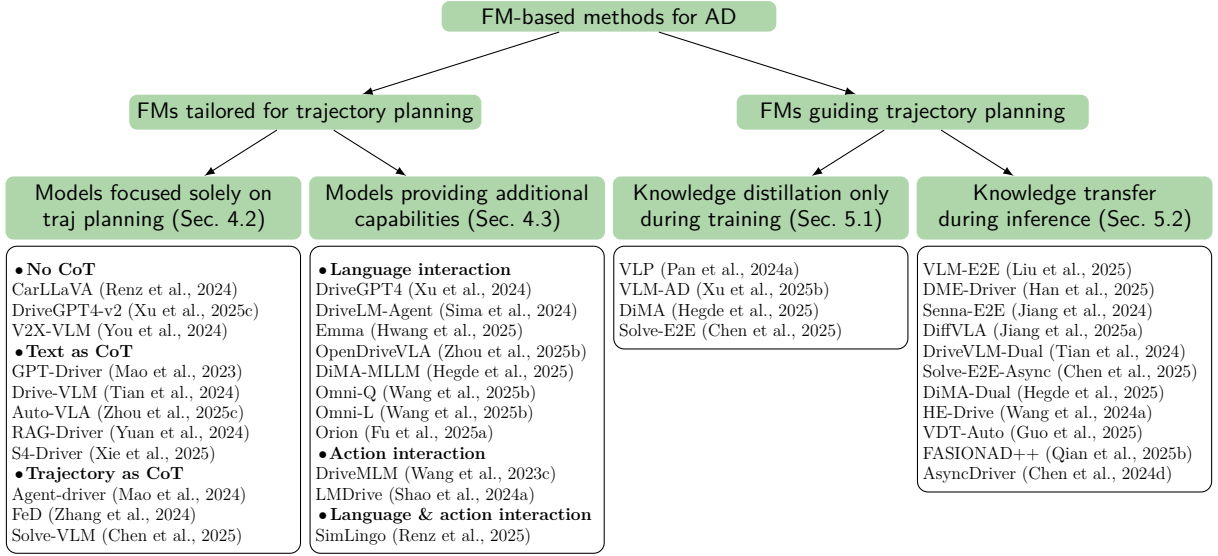

We now present a taxonomy to systematically study trajectory planning approaches utilising \glspl{FM}. Broadly speaking, there are two main categories: \glspl{FM} \textit{tailored for} trajectory planning, and \glspl{FM} \textit{guiding} trajectory planning. 
However, these main categories do involve various subcategories, eventually forming a hierarchical taxonomy as shown in \cref{fig:Taxonomy}. Specifically, each of the eight leaf nodes in the taxonomy tree corresponds to the formulations in \cref{fig:taxonomy_fig}. In what follows, we discuss them in detail.

\subsection[Foundation Models Tailored for Trajectory Planning]{Foundation Models Tailored for Trajectory Planning} \label{subsec:fmtailored}
The primary characteristic of the methods in this group is that they, partly or fully, utilise existing pre-trained \glspl{FM} by tailoring and fine-tuning them to trajectory planning for \gls{AD}. Therefore, effectively, they build \glspl{FM} that are directly used for \gls{AD} use cases. Considering the auto-regressive generation of \glspl{FM} and the convention that the trajectory $\Prediction$ is commonly predicted after the text output $\textoutput$, these approaches can be formulated as:
\begin{align}
    \label{eq:FMforAD}
    \textoutput
    \sim 
    p(\textoutputRandvar | \Observation, \textinput) =  
    \textrm{f}(\Observation, \textinput) \text{, and }     \Prediction
    \sim 
    p(\PredictionRandvar | \Observation, \textinput, \textoutput) =  
    \textrm{f}(\Observation, \textinput, \textoutput),
\end{align}
where $\textrm{f}(\cdot)$ is the fine-tuned \gls{FM}. By design, these methods have the advantage of directly exploiting the vast world knowledge of \glspl{FM} and tailor it towards autonomous driving applications via fine-tuning, a significant advantage over traditional methods for trajectory planning. Additionally, the choice of \gls{FM} also allows these models to have new capabilities. For example, in the case of language-based \glspl{FM}, capabilities such as language- and action-based interactions with users are possible---along with the original task of trajectory planning---during inference. Further details regarding the design choices and categorisation of these methods are provided in \cref{sec:builton}.

We identify 22 current approaches from the literature falling under this broader category. Below, we divide them further into two subgroups depending on the features they have.

\begin{figure*}[t]
        \centering
        \includegraphics[width=0.95\textwidth]{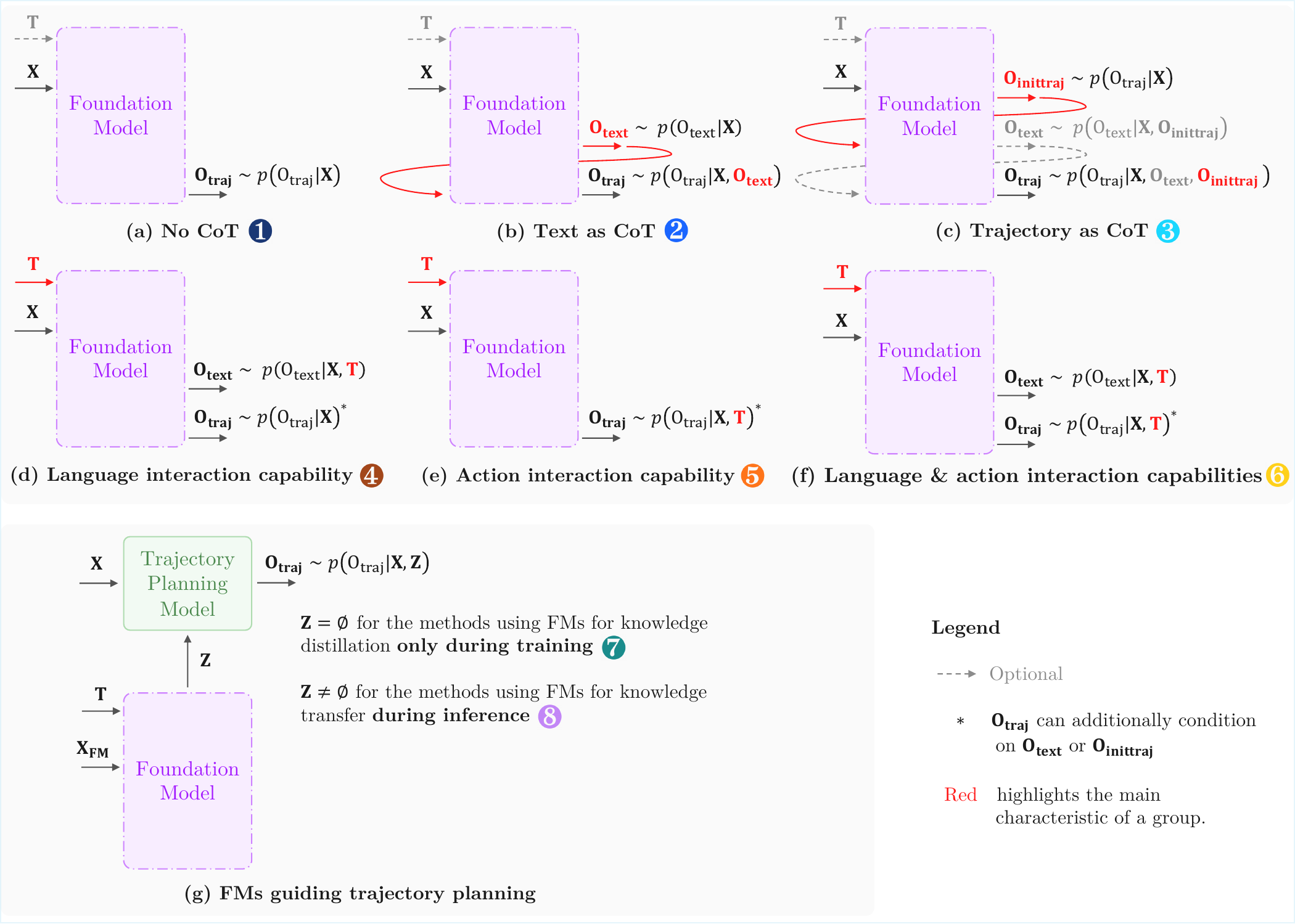}
        \caption{Formulations of subcategories in our taxonomy. \textbf{(a--f)} \glspl{FM} \textit{tailored} for trajectory planning. Specifically, 
        \textbf{(a--c)} are focused solely on trajectory planning. These methods either do not have a text prompt $\textinput$ or have a fixed one, hence $\textinput$ is optional. 
        (c) shows a case where $\Prediction$ optionally conditions on the text output $\textoutput$. Similarly, $\InitialPrediction$ can also benefit from a text output as CoT, which is omitted for clarity.
        \textbf{(d--f)} are the subcategories providing additional capabilities. In (d--f), $\Prediction$ is shown without \gls{CoT} for clarity and $^*$ highlights that $\Prediction$ can also be obtained using different forms of \gls{CoT} in (b) or (c).  
        \textbf{(g)} \glspl{FM} \textit{guiding} trajectory planning via knowledge distillation. 
        }
        \label{fig:taxonomy_fig}
\end{figure*}

\textbf{Models focused solely on trajectory prediction}. These models either have no text prompt $\textinput=\emptyset$, or a fixed one. In the latter case, $\textinput$ simply aims to trigger the model to yield the desired output, e.g., the fixed prompt could be ``\textit{plan the trajectory}''. Since there is no variability in the language, $\Prediction$ primarily relies on input observations $\Observation$.
Methods using \gls{CoT} reasoning solely to provide better trajectories during inference also fall in this category. Similarly, these methods either use a \textit{fixed} single prompt or a set of multiple sequential \textit{fixed} prompts in a pre-defined order to exploit reasoning abilities of \glspl{FM}.
However, the choice of the nature of these prompts plays a crucial role in understanding different factors influencing the decision making of these models, therefore, we further divide this category into three subcategories depending on how, if, \gls{CoT} is used. 
\begin{compactitem}
        \item \Mone~No \gls{CoT}, hence $\textoutput = \emptyset$ and $\Prediction \sim p(\PredictionRandvar|\Observation)$ (see~\cref{fig:taxonomy_fig}(a)). \cref{fig:howfmshelp}(a) sets an example for this category. As an example, given front view image, speed of the ego and GPS target points, CarLLaVA~\citep{CarLLaVA} only outputs the trajectory.
        \item \Mtwo~Text outputs for \gls{CoT}, hence $\textoutput \sim p(\textoutputRandvar | \Observation)$ and $\Prediction \sim p(\PredictionRandvar|\Observation, \textoutput)$ (see~\cref{fig:taxonomy_fig}(b)). 
        \cref{fig:howfmshelp}(b) can be considered as an example of this category.  One example of this subcategory is DriveVLM~\citep{drivevlm}, where a text output including a scene description (e.g., weather, road type, critical objects), influence of the critical objects on ego and planning decision are generated before the model outputs the trajectory.
        \item \Mthree~Initial trajectory prediction $\InitialPrediction$ for \gls{CoT}, hence predicting $\Prediction \sim p(\PredictionRandvar|\Observation, \textoutput, \InitialPrediction)$ (refer~\cref{fig:taxonomy_fig}(c)). Such models first yield $\InitialPrediction$ and then refine it further to $\Prediction$. These models can still leverage text output as well for \gls{CoT} reasoning while generating either $\InitialPrediction$ or $\Prediction$. To illustrate,  FeD~\citep{feedbackguided} aims to improve the initially-predicted trajectory based on the textual feedback it generates, e.g., if $\InitialPrediction$ could result in a collision.
\end{compactitem}

\textbf{Models providing additional capabilities.} In addition to trajectory planning, these models exploit the language understanding of \glspl{FM} to build additional features to cater for  language- and/or action-based interactions. 
Below we further divide them into three subcategories. 
    \begin{compactitem}    
        \item \Mfour~Language interaction capabilities via $\textoutput \sim p(\textoutputRandvar | \Observation,\textinput)$ (see~\cref{fig:taxonomy_fig}(d)). These language interaction capabilities primarily involve question-answer or description tasks, generally for the purpose of informing users about the surrounding or the potential action taken by the model, as also illustrated in \cref{fig:howfmshelp}(c). These capabilities also serve as a tool to understand the decision-making process of these models---allowing practitioners to provide explanations towards model's actions, debug model's outcomes, or to potentially reassure users about its behaviour.
        As an example, Orion~\citep{ORION} can respond to different questions such as ``Could you describe the overall environment and objects captured in the images provided?'' and ``Has the traffic light influenced the driving strategy of the ego vehicle in the previous frames?''.
        Accordingly, the set of the text inputs $\textinput$ that such models can respond is richer than the models providing text output as \gls{CoT}, in which $\textinput$ typically does not exist or it can be limited to a single prompt, e.g., \textit{``What is my future trajectory?''}~\citep{S4Driver}.
        The trajectory planning $\Prediction$ in these models can follow one of the alternatives in \cref{fig:taxonomy_fig}(a-c) depending on if \gls{CoT} reasoning is used.
        \item \Mfive~Action interaction capabilities where text prompt $\textinput$ may directly result in modifications to the planned trajectory by the model (see \cref{fig:taxonomy_fig}(e)). 
        For example, an input that prompts a model to \textbf{follow instructions} would be \textit{``turn left from the next intersection''} and this prompt may result in replanning of the predicted trajectory. Such instruction following prompts indicate the model to behave in a particular way and are normally diverse in nature. Few examples from LMDrive~\citep{lmdrive} include \textit{``Feel free to start driving.''}, \textit{``Depart at the second exit on the roundabout.''}, and \textit{``Execute a right maneuver, prepare for highway exit''}. These are much more diverse compared to the navigational instructions (or route indicators) which consist of pre-defined set of basic commands such as \textit{turn right} or \textit{go straight}, therefore, we do not consider models that can only interact with a navigator in this category. Note, in addition to instructions that directly impact actions, other forms of instructions can surely have indirect impact. For example, an input \textit{``watch out for crossing pedestrians''} that provides a warning would be considered as a \textbf{notice instruction} prompt, while \textit{``crash into the vehicle front''} would be considered as a \textbf{misleading instruction} prompt where the model is expected to avoid following this prompt and follow a safe trajectory. %
    \item \Msix~Some models can also have both action and language interaction capabilities, as illustrated in \cref{fig:taxonomy_fig}(f). SimLingo~\citep{SimLingo} is the only example falling under this subcategory among the methods we consider in this paper.
    \end{compactitem}

\subsection[Foundation Models Guiding Trajectory Planning]{Foundation Models Guiding Trajectory Planning}
Inspired by the well-known knowledge distillation work~\citep{hinton2014distillingknowledgeneuralnetwork}, these models do not build an \gls{FM} for driving use cases, rather they mostly use off-the-shelf \glspl{FM} to help improve their existing trajectory planning models for \gls{AD}. 
From a broader perspective, the methods falling under this category can be formulated as,
\begin{align}
    \label{eq:fmshelpingad}
    \Prediction \sim 
    p(\PredictionRandvar | \Observation, \LatentRepresentation) =  
    \mathrm{f}(\Observation,\LatentRepresentation),
\end{align}
where $\LatentRepresentation$ is the transferred knowledge from the \gls{FM} to the corresponding \gls{AD} model $\mathrm{f}(\cdot,\cdot)$. Note, in this case, $\mathrm{f}(\cdot,\cdot)$ is either a modular approach~\citep{chen2024asynchronous} (see~\cref{fig:adtypes}(a)), where the \gls{FM} can be used at any level to improve trajectory planning, or an \gls{E2E}-\gls{AD} model (see~\cref{fig:adtypes}(b))~\citep{uniad, vad}.

We identify 15 methods in this category and further split them into two subcategories depending on whether $\LatentRepresentation$ is needed during inference or not, as illustrated in \cref{fig:taxonomy_fig}(g).

\textbf{\Mseven~Knowledge distillation only during training.} 
These approaches employ an \gls{FM} for knowledge distillation during training the \gls{AD} model.
As an example, this can be achieved by prompting a \gls{VLM} with a text input and sensor data to obtain a structured output such as a meta-action, which can be distilled into the \gls{AD} model by appending a meta-action prediction module to it, similar to VLM-AD~\citep{VLMAD}.
In such a case the \gls{FM} is not needed for inference, effectively corresponding to $\LatentRepresentation = \emptyset$ in \cref{eq:fmshelpingad}. Hence, the prediction of the model is conditioned only on the observations 
$\Observation$ during inference, i.e., $\Prediction \sim p(\PredictionRandvar | \Observation)$. 
This offers the advantage of maintaining the inference efficiency of the \gls{AD} model as the \gls{FM} is not needed for inference.

\textbf{\Meight~Knowledge transfer during inference.} 
These methods utilise an \gls{FM} not only during training, but also during inference with the intention to leverage the knowledge of \glspl{FM} more effectively. This corresponds to $\LatentRepresentation \neq \emptyset$ in \cref{eq:fmshelpingad}, where
$\LatentRepresentation$ is usually taken as \textit{a scene description}~\citep{vlme2e}, typically involving perception knowledge such as the objects in the scene, or \textit{a planning decision}, which can include a trajectory~\citep{drivevlm,solve} or a meta-action~\citep{Senna} from the \gls{FM}. In either of the cases, $\LatentRepresentation$ is typically used either as an internal representation of the \gls{FM}, or directly as the output of the \gls{FM}. In the latter, the \gls{FM} output can also be a plain text, in which case an additional text encoder is typically employed for encoding the text before propagating to the \gls{AD} model. Nevertheless, due to the use of \gls{FM} at inference, this group of approaches typically requires additional compute for inference compared to the methods mentioned in the category above. 
\section{Foundation Models Tailored for Trajectory Planning}
\label{sec:builton}
We now delve deeper into the design and development of \glspl{FM} tailored for trajectory planning for \gls{AD}. 
We first elaborate on the important ingredients one must pay attention to before even beginning to fine-tune an \gls{FM} for trajectory planning in \cref{subsec:recipe}, and then discuss different approaches for each of the subcategories in detail in \cref{subsec:trajplanningonly} and \cref{subsec:additionalcapabilities}.

\begin{figure*}[t]
        \centering
        \includegraphics[width=1.0\textwidth]{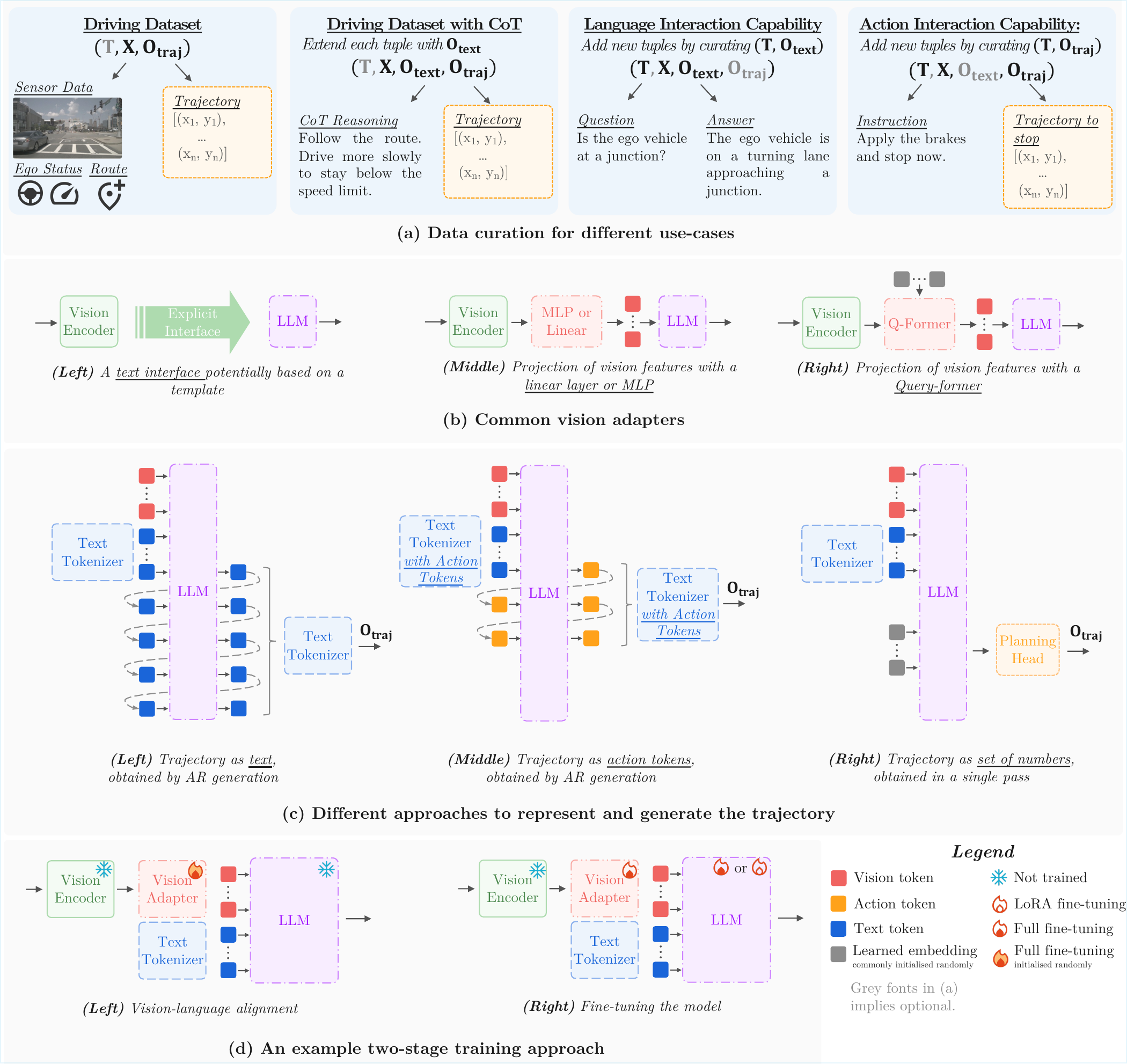}
        \caption{The steps to fine-tune an \gls{FM} for trajectory planning. \textbf{(a)} \textbf{Data curation.} \textbf{(b)} For the \textbf{model design}, either an off-the-shelf \gls{VLM} can be used or a vision encoder can be combined with an \gls{LLM} using one of the depicted vision adapters. \textbf{(c)} \textbf{Trajectory representation.} Vision encoder and adapter are omitted for clarity. \textbf{(d)} Among various \textbf{model training strategies}, this is an example two-stage training approach based on \cite{LLaVA}, usually adopted when an off-the-shelf \gls{VLM} is \textbf{not} used.}
        \label{fig:designdecisions}
\end{figure*}

\subsection[How to Fine-tune a FM for Trajectory Planning]{How to Fine-tune an \gls{FM} for Trajectory Planning}
\label{subsec:recipe}

\textbf{Data Curation.} The structure of the data to be curated while adapting an \gls{FM} for trajectory planning depends primarily on the desired use case of the model, as illustrated in \cref{fig:designdecisions}(a). 
\begin{compactitem}
    \item \textit{Driving Dataset.} The dataset for trajectory planning would simply comprise the pairs of observations and the trajectory, i.e., $(\Observation, \Prediction)$. As aforementioned, the observations ($\Observation$) typically include a subset of the sensor data, the ego status, and an indicator of the route. An example of this form of dataset is nuScenes~\citep{nuscenes}, including multi-view camera, radar and lidar data as sensor input, and future ego positions as the trajectory target. 
    \item \textit{Driving Dataset with CoT Reasoning.} For complex situations where reasoning is crucial, one might be interested in exploiting the \gls{CoT} reasoning ability of \glspl{FM} for enhanced understanding of the scene. To enable that, each tuple in the driving dataset is extended by $\textoutput$ (the text to enforce \gls{CoT} reasoning) leading to the dataset consisting of tuples $(\Observation, \textoutput, \Prediction)$. An example dataset with textual descriptions that can be used as \gls{CoT} reasoning is BDD-X~\citep{bddx}.
    When CoT is performed using an initial trajectory $\InitialPrediction$, a dataset might not be curated and stored in advance, and instead it is generated by the model. These driving datasets could also include a typically fixed input prompt $\textinput$, such as the example in \cref{fig:howfmshelp}(b) with $\textinput=$\textit{``Plan the trajectory''}. 
    \item \textit{Language Interaction Capability.} Although this resulting driving dataset can be used for trajectory planning, additional tuples are needed to account for the language or action interaction capabilities, depending on the use case. Specifically, for the language interaction capability, the dataset is  extended with new tuples, $(\Observation, \textinput, \textoutput)$ where $\textinput$, $\textoutput$ is the question-answer pair grounding on $\Observation$. An example dataset with language interaction is the Chat-B2D dataset used by Orion~\citep{ORION}.
    \item \textit{Action Interaction Capability.} If the model is expected to generate a trajectory, while answering the question, then such tuples include $\Prediction$, i.e., $(\Observation, \textinput, \textoutput, \Prediction)$, in which case the answer $\textoutput$ could serve as \gls{CoT} reasoning. As for the action interaction capability, the model is expected to consider the user instruction for trajectory planning. Accordingly, the data tuples typically follow $(\Observation, \textinput, \Prediction)$, such that $\textinput$ is the user instruction and $\Prediction$ is the trajectory corresponding to $\textinput$. An example dataset with language and action interaction is the SimLingo dataset~\citep{SimLingo}.
    Again, optionally, $\textoutput$ can be included for \gls{CoT} reasoning during instruction following. We further discuss the different choices of the \gls{CoT} reasoning in \cref{subsec:trajplanningonly} and the datasets designed to include additional capabilities in \cref{subsec:additionalcapabilities}. 
\end{compactitem}

\textbf{Model Design.} While designing the model architecture, some approaches~\citep{SimLingo,autovla} use off-the-shelf \glspl{VLM}, providing the advantage of initialising all parameters jointly from a state trained on a large dataset. Differently, another group of approaches~\citep{drivegpt4,solve,ORION} combines a preferred vision encoder, such as ViT~\citep{vit}, with a preferred \gls{LLM} using a randomly-initialised vision adapter. Both approaches follow the typical \gls{VLM} architecture (see \cref{fig:adtypes}(c)), where the vision adapter connects the vision encoder and the \gls{LLM}. Additionally, the latter approach to combine preferred vision encoder and \gls{LLM}, requires designing a vision adapter. As shown in \cref{fig:designdecisions}(b), in general, there are three different choices of the vision adapter:
\begin{compactitem}
    \item The vision adapter can be a \textit{text interface} where the vision encoder directly outputs perception and/or prediction information such as the locations of the objects and map elements in the scene. This information is then propagated to the \gls{LLM} as text, in which case the model is not trained end-to-end as in \cite{gptdriver}. 
    \item  \textit{A linear layer or an \gls{MLP}} can project all vision tokens from the vision encoder onto the \gls{LLM} input space~\citep{LLaVA} as used by \cite{SimLingo}.
    \item Different from a linear layer or \gls{MLP} block, \textit{Queryformer}~\citep{blip2} relies on cross-attention between a set of typically randomly-initialised latent representations and the tokens from the vision encoder to selectively project the most useful set of representations. This approach is taken by Omni-Q~\citep{omnidrive}.
\end{compactitem}
Additionally, including custom modules can impose certain inductive biases absent in off-the-shelf \glspl{FM}. For example, as off-the-shelf \glspl{VLM} are generally not pretrained on 3D perception tasks or using videos, some approaches design custom modules~\citep{ORION, omnidrive, S4Driver, solve} to enforce the model to consider these clues absent in the \glspl{VLM}.

\textbf{Trajectory Representation.}
Another crucial aspect in tailoring an \gls{FM} for \gls{AD} is how to represent the trajectory, as this can consequently require modifications in the model design as well. \cref{fig:designdecisions}(c) illustrates common trajectory representations, which we elaborate on below:
\begin{compactitem}
    \item \textit{Trajectory as text, obtained by the standard \gls{AR} text generation of the \gls{LLM}.} For example, consider a model that generates a single character in each pass and assume that $1.54, 0.21$ is a 2D point of the trajectory. As a result, this 2D point is generated sequentially as `1', `.', `5', `4', `,', `0', `.', `2', `1' in an \gls{AR} manner, as illustrated in \cref{fig:designdecisions}(c)(Left).  Consequently, while this approach generally does not require any modification to the tokenizer or model architecture, it requires multiple passes through the \gls{LLM} to yield a single point of the trajectory, increasing the inference time.
    \item \textit{Trajectory as action tokens.} Instead of generating a single 2D point in multiple passes, one can discretise the action space and represent these discrete actions in the vocabulary of the \gls{LLM} by the action tokens. One example is to discretise 2D \gls{BEV} space using a grid and use the points on the grid as the action tokens~\citep{drivelm,rt2}. These action tokens can be included in the tokenizer by mapping the rarely-used tokens in the vocabulary to each of these 2D points~\citep{rt2}. \cite{autovla} also follow a similar approach by building a vocabulary for the actions with 2048 actions determined by clustering the actions in terms of the relative 2D position and the heading angle in the next $0.5$ seconds. In these examples, the trajectory consists of multiple points, and hence multiple \gls{AR} passes within the \gls{LLM} are required to obtain a trajectory, as shown in \cref{fig:designdecisions}(c)(Middle). As an alternative, similar to the planning vocabulary in VADv2~\citep{vadv2}, an action token can represent an entire trajectory where the trajectory can be obtained in a single pass, however, this might be more prone to errors due to long-horizon planning.
    \item \textit{Trajectory as a set of numbers, generally obtained in a single pass by using a planning head.} Alternatively, the trajectory can be obtained by an additional planning head that utilises the representations learned by the \gls{LLM}. There are two main variants of this approach. As shown in \cref{fig:designdecisions}(c)(Right), some methods~\citep{CarLLaVA,feedbackguided,SimLingo} use learned embeddings that attend to the observations (and a text input, if exists) before being decoded as a trajectory by the planning head. On the other hand, another group of approaches~\citep{ORION,DriveGPT4v2} does not introduce learned embeddings, and instead the planning head relies on the representations of the \gls{LLM} in the last layer. In any case, this approach requires designing a planning head to be appended to the \gls{LLM}. \gls{MLP}~\citep{SimLingo} and generative models, such as a variational autoencoder~\citep{ORION} or diffusion model~\citep{DiffusionDrive}, have been explored in the literature for this purpose.
\end{compactitem}

\begin{table*}[t]
    \setlength{\tabcolsep}{0.2em}
    \centering
    \caption{Design choices of \glspl{FM} tailored for \gls{AD} (refer to \cref{fig:Taxonomy}).  
    Unless mentioned otherwise in the corresponding paper, we assume that each method uses full parameter fine-tuning of a module, generates the trajectory in the form of text, and does not have route indicator as an input. 
    Custom \gls{VLM} implies that the model does not employ an off-the-shelf \gls{VLM} but combines a vision encoder with an \gls{LLM} itself, in which case we specify the details of the vision encoder and \gls{LLM}, e.g., SigLIP-ViT-G as in the case of S4-Driver.
    Please refer to \cref{sec:builton} for the details of the training symbols.}
    \label{tab:trajectoryplanningmethods}
    \scalebox{0.54}{
\begin{tabular}{
>{\centering\arraybackslash\hspace{0pt}}m{0.025\linewidth}
>{\centering\arraybackslash\hspace{0pt}}m{0.025\linewidth}
>{\hspace{0pt}}
>{\arraybackslash}m{0.20\linewidth}
>{\centering\arraybackslash}m{0.15\linewidth}
>{\centering\arraybackslash}m{0.10\linewidth}
>{\centering\arraybackslash}m{0.15\linewidth}
>{\centering\arraybackslash}m{0.30\linewidth} 
>{\centering\arraybackslash}m{0.20\linewidth}
>{\centering\arraybackslash}m{0.28\linewidth}
>{\centering\arraybackslash}m{0.11\linewidth}
>{\centering\arraybackslash}m{0.18\linewidth}}
    \toprule
    &&&\multicolumn{2}{c}{\textbf{Observations}}&\multicolumn{5}{c}{\textbf{Model Design and Model Training}}&\\
    \cmidrule(lr){4-5}
    \cmidrule(lr){6-10}\\
    &&\textbf{Model}&\textbf{Sensors}&\textbf{Route indicator}&\textbf{\gls{VLM}}&\textbf{Vision Encoder}&\textbf{Vision Adapter}&\textbf{\gls{LLM}}&\textbf{\# Params}&\textbf{Trajectory Representation}\\ \midrule
    \multirow{5}{*}{\Mone}&1&\textit{CarLLaVA}\par{}\citep{CarLLaVA}&Front camera&\cmark&LLaVA-Next&ViT-L\includegraphics[height=0.02\textheight]{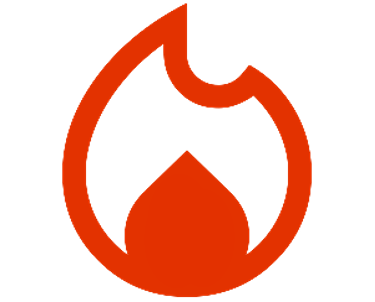}&Linear\includegraphics[height=0.02\textheight]{Images/Finetune.png}&Tiny-Llama\includegraphics[height=0.02\textheight]{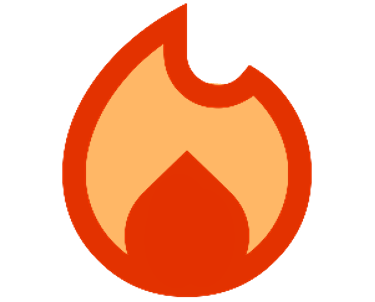}&>350M&Numbers\\
    &2&\textit{DriveGPT4-v2}\par{}\citep{DriveGPT4v2}&3 Front cameras&\cmark&Custom&SigLIP-ViT-L\includegraphics[height=0.02\textheight]{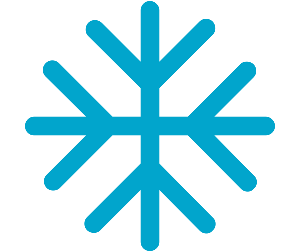}&Linear\includegraphics[height=0.02\textheight]{Images/FullTrain.png}&Qwen-0.5B\includegraphics[height=0.02\textheight]{Images/Finetune.png}&>1B&Numbers\\
    &3&\textit{V2X-VLM}\par{}\citep{v2xvlm}&Front+Infra camera&\xmark&Florence-2&DaViT\includegraphics[height=0.02\textheight]{Images/Frozen.png}&Linear\includegraphics[height=0.02\textheight]{Images/Finetune.png}&BART\includegraphics[height=0.02\textheight]{Images/Finetune.png}&232B&Text\\
    \midrule
    \multirow{9}{*}{\Mtwo}&4&\textit{GPT-Driver}\par{}\citep{gptdriver}&360$^{\circ}$ cameras&\cmark&Custom&Based on UniAD\includegraphics[height=0.02\textheight]{Images/Frozen.png} &Text&Chat-GPT3.5\includegraphics[height=0.02\textheight]{Images/Finetune.png}&>175B&Text\\ 
    &5&\textit{Drive-VLM}\par{}\citep{drivevlm}&Front camera&\xmark&Qwen-VL&Custom ViT\includegraphics[height=0.02\textheight]{Images/Finetune.png}&\gls{MLP}\includegraphics[height=0.02\textheight]{Images/Finetune.png}&QwenLM\includegraphics[height=0.02\textheight]{Images/Finetune.png}&9.6B&Text\\
    &6&\textit{Auto-VLA}\par{}\citep{autovla}&3 Front cameras&\cmark&Qwen2.5-VL&Custom ViT\includegraphics[height=0.02\textheight]{Images/Finetune.png}&\gls{MLP}\includegraphics[height=0.02\textheight]{Images/Finetune.png}&Qwen2.5\includegraphics[height=0.02\textheight]{Images/Finetune.png}&>3B&Action token\\
    &7&\textit{RAG-Driver}\par{}\citep{ragdriver}&Front camera&\xmark&Video-LLaVA&ViT-B\includegraphics[height=0.02\textheight]{Images/Frozen.png}&Linear\includegraphics[height=0.02\textheight]{Images/Finetune.png}&LLaMA2\includegraphics[height=0.02\textheight]{Images/Finetune.png}&>7B&Text\\
    &8&\textit{S4-Driver}\par{}\citep{S4Driver}&360$^{\circ}$ cameras&\cmark&P\removal{A}\addition{a}LI-3&SigLIP-ViT-G\includegraphics[height=0.02\textheight]{Images/Frozen.png}+Custom\includegraphics[height=0.02\textheight]{Images/FullTrain.png}&Linear\includegraphics[height=0.02\textheight]{Images/Finetune.png}&UL2\includegraphics[height=0.02\textheight]{Images/Finetune.png}&>5B&Text\\
    \midrule
    \multirow{5}{*}{\Mthree}&9&\textit{Agent-Driver}\par{}\citep{agentdriver}&360$^{\circ}$ cameras&\cmark&Custom&Based on UniAD\includegraphics[height=0.02\textheight]{Images/Frozen.png} &Text&Chat-GPT3.5\includegraphics[height=0.02\textheight]{Images/Finetune.png}&>175B $\times$ 2&Text \\
    &10&\textit{FeD}\par{}\citep{feedbackguided}&Front camera&\cmark&LLaVA&ViT-L\includegraphics[height=0.02\textheight]{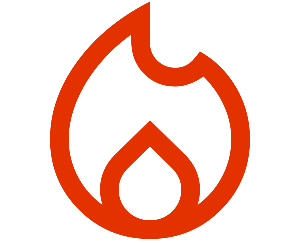}&Linear\includegraphics[height=0.02\textheight]{Images/LoRA.png}&Llama\includegraphics[height=0.02\textheight]{Images/LoRA.png}&>7B&Numbers\\
    &11&\textit{Solve-VLM}\par{}\citep{solve}&360$^{\circ}$ cameras&\xmark&Custom&EVA-02-L\includegraphics[height=0.02\textheight]{Images/Finetune.png}&Q-Former\includegraphics[height=0.02\textheight]{Images/FullTrain.png}&LLaVA-v1.5-LLM\includegraphics[height=0.02\textheight]{Images/LoRA.png}&>7B&Text\\
    \midrule
    \multirow{15}{*}{\Mfour}&12&\textit{DriveGPT4}\par{}\citep{drivegpt4}&Front camera&\xmark&Custom&Valley-ViT-L\includegraphics[height=0.02\textheight]{Images/Frozen.png}& Valley-Custom\includegraphics[height=0.02\textheight]{Images/FullTrain.png}&Llama2\includegraphics[height=0.02\textheight]{Images/Finetune.png}&>7B&Text\\
    &13&\textit{DriveLM-Agent}\par{}\citep{drivelm}&Front camera&\xmark&BLIP-2&ViT-L\includegraphics[height=0.02\textheight]{Images/LoRA.png}&Q-Former\includegraphics[height=0.02\textheight]{Images/LoRA.png}&Flan-T5\includegraphics[height=0.02\textheight]{Images/LoRA.png}&>7B&Action token\\
    &14&\textit{Emma}\par{}\citep{emma}&360$^{\circ}$ cameras&\cmark&Gemini&Unknown\includegraphics[height=0.02\textheight]{Images/Finetune.png}&Unknown\includegraphics[height=0.02\textheight]{Images/Finetune.png}&Unknown\includegraphics[height=0.02\textheight]{Images/Finetune.png}& 1.8B&Text\\
    &15&\textit{OpenDriveVLA}\par{}\citep{OpenDriveVLA}&360$^{\circ}$ cameras&\cmark&Custom&Bevformer\includegraphics[height=0.02\textheight]{Images/FullTrain.png}&Q-Former+MLP\includegraphics[height=0.02\textheight]{Images/FullTrain.png}&Qwen 2.5-Instruct\includegraphics[height=0.02\textheight]{Images/Finetune.png}&>0.5B-7B&AR text\\
    &16&\textit{DiMA-MLLM}\par{}\citep{dima}&360$^{\circ}$ cameras&\xmark&Custom&Bevformer\includegraphics[height=0.02\textheight]{Images/FullTrain.png}&Q-Former\includegraphics[height=0.02\textheight]{Images/FullTrain.png}&LLaVA-v1.5-LLM\includegraphics[height=0.02\textheight]{Images/LoRA.png}&>7B&Text\\
    &17&\textit{Omni-L}\par{}\citep{omnidrive}&360$^{\circ}$ cameras&\cmark&Custom&EVA-02-L\includegraphics[height=0.02\textheight]{Images/Finetune.png}&\gls{MLP}\includegraphics[height=0.02\textheight]{Images/Finetune.png}&LLaVA-1.5-LLM\includegraphics[height=0.02\textheight]{Images/Finetune.png}&>7B&Text\\
    &18&\textit{Omni-Q}\par{}\citep{omnidrive}&360$^{\circ}$ cameras&\cmark&Custom&EVA-02-L\includegraphics[height=0.02\textheight]{Images/Finetune.png}&Q-Former\includegraphics[height=0.02\textheight]{Images/FullTrain.png}&LLaVA-1.5-LLM\includegraphics[height=0.02\textheight]{Images/Finetune.png}&>7B&Text\\
    &19&\textit{Orion}\par{}\citep{ORION}&360$^{\circ}$ cameras&\cmark&Custom&EVA-02-L\includegraphics[height=0.02\textheight]{Images/Finetune.png}&Q-Former\includegraphics[height=0.02\textheight]{Images/FullTrain.png}&Vicuna-1.5\includegraphics[height=0.02\textheight]{Images/LoRA.png}&>7B&Numbers\\
    \midrule
    \multirow{3}{*}{\Mfive}&20&\textit{DriveMLM}\par{}\citep{drivemlm}&360$^{\circ}$ cameras, lidar&\cmark&Custom&EVA-ViT-G\includegraphics[height=0.02\textheight]{Images/Frozen.png}+GD-MAE\includegraphics[height=0.02\textheight]{Images/Frozen.png}&Q-Former\includegraphics[height=0.02\textheight]{Images/FullTrain.png}&Llama\includegraphics[height=0.02\textheight]{Images/Finetune.png}&>7B&Text\\
    &21&\textit{LMDrive}\par{}\citep{lmdrive}&360$^{\circ}$ cameras, lidar&\cmark&Custom&ResNet\includegraphics[height=0.02\textheight]{Images/Frozen.png}+PointPillars\includegraphics[height=0.02\textheight]{Images/Frozen.png}&Q-Former\includegraphics[height=0.02\textheight]{Images/FullTrain.png}&LLaVA-v1.5-LLM\includegraphics[height=0.02\textheight]{Images/Frozen.png}&>7B&Numbers\\
    \midrule
    \Msix&22&\textit{SimLingo}\par{}\citep{SimLingo}&Front camera&\cmark&Mini-InternVL&InternViT\includegraphics[height=0.02\textheight]{Images/Finetune.png}&\gls{MLP}\includegraphics[height=0.02\textheight]{Images/Finetune.png}&Qwen2\includegraphics[height=0.02\textheight]{Images/LoRA.png}&1B&Numbers\\
    \bottomrule
    \end{tabular}
     }
\end{table*}

\textbf{Model Training.} Fine-tuning the \gls{FM} is typically carried out in a single step or in multiple steps, where each step generally targets a specific subset of the modules. For example, similar to LLaVA-style training~\citep{LLaVA}, as illustrated in \cref{fig:designdecisions}(d), one can first train the vision adapter by freezing the vision encoder and \gls{LLM}, especially when a custom vision adapter is introduced with randomly-initialised parameters. After training the adapter, all model parameters or a subset of them can be fine-tuned for the target task. To provide an overview of how the existing methods in the literature fine-tune their models, we introduce the following symbols that represent the change of the parameters in a module across training (spanning all stages in the case of multi-stage training): 

\begin{compactitem}
    \item \includegraphics[height=0.015\textheight]{Images/Frozen.png} indicates keeping modules completely frozen during training.
    \item \includegraphics[height=0.015\textheight]{Images/LoRA.png} represents \gls{LoRA}-style fine-tuning of a module~\citep{lora}, which assumes that the adaptation needed for new tasks can be captured by low-rank updates. This approach is efficient but limits the capacity of the model for adaptation.
    \item \includegraphics[height=0.015\textheight]{Images/Finetune.png} implies the standard fine-tuning of all parameters, utilising the full capacity of the module for fine-tuning, though it requires more resources than using \gls{LoRA}.
    \item \includegraphics[height=0.015\textheight]{Images/FullTrain.png} represents training a module by initialising it either randomly, e.g., once a custom module is introduced, or from a model pretrained only on domain-specific data, e.g., a vision encoder trained only on nuScenes~\citep{nuscenes}.
\end{compactitem}

Using these symbols, \cref{tab:trajectoryplanningmethods} presents the main design choices of 22 methods within the scope of this section.

\begin{table*}[t]
    \setlength{\tabcolsep}{0.2em}
    \centering
    \caption{Example models solely focused on trajectory planning. For such models, $\textinput$ is either $\emptyset$ or fixed, and 
    \gls{CoT} reasoning includes $\textoutput$ or an initially-planned trajectory $\InitialPrediction$ for self-correction. 
    \label{tab:trajonlyinputexamples}}
    \scalebox{0.64}{
\begin{tabular}{
>{\arraybackslash}m{0.34\linewidth}
>{\arraybackslash\columncolor{gray!8}}m{0.24\linewidth}
>{\arraybackslash}m{0.34\linewidth}
>{\arraybackslash\columncolor{gray!8}}m{0.24\linewidth}
>{\arraybackslash}m{0.34\linewidth}}
    \toprule
    \centering{\textbf{Model Type}}&\centering{\textbf{Example Model}}&\centering{\textbf{Observations ($\Observation$)}}&\centering{\textbf{System Prompt ($\textinput$)}}&\textbf{Text Output ($\textoutput$)}
    \\ \midrule
    $\Prediction \sim p(\PredictionRandvar|\Observation)$&\textit{CarLLaVA}&Front image; speed of ego; GPS target point&$\emptyset$&$\emptyset$\\ 
    $\Prediction \sim p(\PredictionRandvar|\Observation, \textoutput)$&\textit{S4-Driver}&Multiview images; speed, acceleration and past trajectory of ego; high level command&What is my future trajectory?&Meta-decision, e.g., keep speed then do decelerating\\
    $\Prediction \sim p(\PredictionRandvar|\Observation, \textoutput,\InitialPrediction)$&\textit{FeD}&Front image; speed of ego; GPS target point and high level command&Please evaluate the predicted future locations of ego vehicle&Collision with vehicles or pedestrians, traffic light violations, deviation from the expert and planned route\\
    \bottomrule
    \end{tabular}
     }
\end{table*}

\subsection{Models Focused Solely on Trajectory Planning}
\label{subsec:trajplanningonly}
Following the taxonomy provided in \cref{fig:Taxonomy}, we now discuss different subcategories.
\cref{tab:trajonlyinputexamples} provides an example approach for each of these subcategories.

\subsubsection[Models without CoT Reasoning \Monesymb]{Models without \gls{CoT} Reasoning \Monesymb}
\label{subsubsec:withoutcot}
As one of the earlier approaches that uses a \gls{VLM} for trajectory planning, \textbf{CarLLaVA}~\citep{CarLLaVA} is built on LLaVA-NeXt~\citep{LLaVAnext}, which combines ViT~\citep{vit} vision encoder with Tiny-Llama~\citep{llama} as the \gls{LLM}. Given the front-view image, current ego speed and two GPS target points as a route indicator, the model predicts (i) 20 \textit{path waypoints} that are equidistant points at 1 meter apart, thus independent of the time, to control the steering and (ii) 10 \textit{speed waypoints}, which are points at equal time intervals, specifically 0.25 seconds apart, to control the accelerator and the brake. These outputs are obtained in one-pass using \glspl{MLP} as the planning head, following the design in \cref{fig:designdecisions}(c)(Right). Note, unlike other methods we explore, the Tiny-Llama in CarLLaVA is trained from scratch. In 2024, this model, trained on approximately 3 million images, won the CARLA 2.0 challenge~\citep{carla}, a competition based on driving in a closed-loop simulator. Given three camera views as front, front-left and front-right, \textbf{DriveGPT4-v2}~\citep{DriveGPT4v2}, as a similar model, combines SigLIP-ViT~\citep{siglip} with Qwen~\citep{qwen} to predict the target speed and turning angle controls directly. In addition to that, the model is supervised to predict a trajectory and equidistant \textit{route waypoints}, similar to the path waypoints of CarLLaVA. Different from CarLLaVA, \textit{an expert model provided with privileged information}, such as the objects in the scene and potential hazard information, is trained to augment the training dataset of DriveGPT4-v2. This augmentation is shown to have a notable effect on driving performance. While these two approaches do not have a system prompt $\textinput$, \textbf{V2X-VLM}~\citep{v2xvlm} uses a fixed $\textinput=\textit{``Please predict the ego vehicle positions over next 45 timestamps.''}$. V2X-VLM is built on Florence-2~\citep{florence2} (using DaViT image encoder~\citep{davit} with BART language model~\citep{lewis2020bart}), and utilises DAIR-V2X dataset~\citep{yu2022dair} to focus specifically on \textit{leveraging an external camera} from the infrastructure in the environment.

\begin{observation}{Models without \gls{CoT} Reasoning \Mone}
As these approaches are solely designed for driving, the dataset curation is relatively simple, where each data example typically consists of $(\Observation, \Prediction)$ pairs. Moreover, due to the absence of \gls{CoT} reasoning, these models demonstrate superior computational efficiency particularly when trajectory generation can be accomplished through a single forward pass. Nevertheless, these advantages entail certain trade-offs: the omission of \gls{CoT} reasoning may limit driving performance, and the lack of language generation capabilities diminishes their explainability.
\end{observation}

\subsubsection[Text Output for CoT Reasoning \Mtwosymb]{Text Output for CoT Reasoning \Mtwosymb
}
\label{subsubsec:textoutputcot}
\gls{CoT} reasoning is typically a prompting strategy of the \glspl{FM} that help them to break down the problem into multiple steps during inference. This is commonly achieved by prepending a step-by-step solution to a similar problem~\citep{chainofthought,treeofthoughts,algorithmofthoughts} to the input prompt, as a demonstration on how to approach and solve the problem. Furthermore, simply appending  \textit{``Let’s think step by step''} to the prompt is also found to be quite effective, also known as the \textit{zero-shot \gls{CoT}}~\citep{zeroshotcot}. These enhanced ways of prompting strategies facilitate \glspl{FM} to yield its output step-by-step, improving their performance in certain tasks such as arithmetic reasoning.

Similarly, \glspl{FM} tailored for trajectory planning leverage \gls{CoT} reasoning for improving the driving performance, however, in a slightly nuanced manner. In particular, these approaches explicitly \textit{fine-tune the \glspl{FM}} to yield the output step-by-step, instead of modifying the input prompt $\textinput$ during inference time. Besides, the prompt $\textinput$, in this case, is typically fixed, e.g., \textit{``What is my future trajectory?''} (see S4-Driver in \cref{tab:trajonlyinputexamples}), and can potentially involve multiple steps. As a result of the fine-tuning, text output $\textoutput$ includes a predefined set of explanations that are expected to improve trajectory planning, i.e., $\Prediction$. The scope of $\textoutput$ varies significantly across methods, ranging from only a meta-action to a comprehensive description of the scene and agents' behaviours (please refer to \cref{fig:howfmshelp}(b) for a more comprehensive example). 

Methods using a more comprehensive $\textoutput$ as the \gls{CoT} reasoning generally follow the common pipeline of the modular \gls{AD} models (see \cref{fig:adtypes}(a)), hence generating the \gls{CoT} reasoning in the order of perception and prediction information followed by planning decision, e.g, meta-action and trajectory. As a pioneering approach following this sequence, \textbf{GPT-Driver}~\citep{gptdriver} fine-tunes Chat-GPT3.5~\citep{chatgpt} to yield notable objects in the scene, along with their potential impact on ego, and a meta-action as \gls{CoT} reasoning before yielding the proposed trajectory. As an earlier method, GPT-Driver adopts a \textit{text interface} as the vision adapter (refer to \cref{fig:designdecisions}(b)(Left)), hence it is not \gls{E2E}-trainable. Specifically, the perception and prediction information is extracted by the corresponding modules of UniAD~\citep{uniad}, and propagated to Chat-GPT3.5 as text. \textbf{DriveVLM}~\citep{drivevlm} follows a similar sequence in terms of \gls{CoT} reasoning, where scene description and analysis are followed by meta-action prediction, and finally trajectory planning. \textit{To preserve the capabilities} that the \gls{VLM} gained before fine-tuning, Qwen-VL~\citep{qwenvl} is fine-tuned by combining \gls{AD} datasets with the LLaVA dataset, which includes examples from different domains, referred to as \textit{co-tuning}. Similar to the previous methods, \textbf{Auto-VLA}~\citep{autovla} also relies on a comprehensive $\textoutput$ for \gls{CoT} reasoning, including a scene description, identification of the crucial objects, the intentions of the surrounding agents, and ideal driving actions. Differently, it follows \textit{an adaptive \gls{CoT} mechanism} where the model refers to \gls{CoT} only for complex scenarios, considering its inefficient nature. To train such a model, a dataset is constructed as a combination of action-only scenes and reasoning-augmented scenes. The model is first trained using supervised fine-tuning, followed by reinforcement learning, specifically GRPO~\citep{grpo}, where the reward function promotes driving-related measures such as safety and comfort while discouraging \gls{CoT} reasoning if deemed unnecessary.

There are also methods that generate a brief description or explanation as \gls{CoT} reasoning. One example is \textbf{RAG-Driver}~\citep{ragdriver}, which only predicts an explanation of the action that the ego is executing, such as \textit{``The car moves forward then comes to a stop because traffic has stopped in front''}, known as action explanation and justification in BDD-X dataset~\citep{bddx}. Relying on this as \gls{CoT} reasoning, the model outputs speed and turning angle of the ego. As the model is named after, it leverages \textit{\gls{RAG}}, which retrieves relevant information from external sources to enhance the model's prediction. Accordingly, in the \gls{AD} domain, given a test sample, the two most similar samples in terms of cosine similarity are retrieved from the training data and fed to the \gls{LLM}. As the retrieved data have known control signals, they are expected to improve driving performance for the test sample. Alternatively, \textbf{S4-Driver}~\citep{S4Driver} show that it is even useful to predict only one of the four pre-determined meta-actions (i.e., \{keep stationary, keep speed, accelerate and decelerate\}) via \gls{CoT} reasoning before planning the trajectory. The model is built on PaLI-3 \gls{VLM}~\citep{pali3}, which is based on the UL2 \gls{LLM}~\citep{tay2022ul2}, and equipped with custom modules to facilitate learning 3D scene representations. 

\textbf{The inference time and accuracy trade-off of \gls{CoT} reasoning.} One key impact of \gls{CoT} reasoning that needs to be taken into consideration is its effect on inference time. 
This is because $\textoutput$, as the \gls{CoT} reasoning, is typically generated in an \gls{AR} manner, requiring multiple passes through the \gls{LLM}.
This is especially important for the applications requiring fast inference speed, such as trajectory planning.
To provide an example, we test the throughput of SimLingo, a relatively lightweight model with 1B parameters that processes only the front image, with and without \gls{CoT} reasoning. 
Specifically for SimLingo, $\textinput=\textit{``Predict the waypoints.''}$ prompt yields the trajectory without \gls{CoT} reasoning, while using $\textinput=\textit{``What should the ego do next?''}$ triggers a relatively lightweight \gls{CoT} reasoning with 2 or 3 short sentences about the what the action should be and why.
For each of these prompts, we propagate 200 images to SimLingo five times using its official implementation. 
%
%
We observe that the \gls{CoT} reasoning increases the inference time of SimLingo by 4.5$\times$, from  3.6fps to 0.8fps on an Nvidia A4500 GPU, which can be a significant bottleneck for practical deployment.
In this case, the authors report a relatively small performance gain of using \gls{CoT} reasoning, where the driving score increased from $84.41$ to $85.07$ and success rate improved from $64.84$ to $67.27$ on Bench2Drive benchmark~\citep{bench2drive}.
As a result, the compute requirement of the designed \gls{CoT} reasoning as well as the performance gain it contributes to should be taken into consideration while using it in practice.

\begin{observation}
{Models using Text Output for \gls{CoT} \Mtwo} While text-based \gls{CoT} reasoning offers the advantage of grounding trajectories in the model's rationale, thereby enhancing explainability and potentially driving performance, these benefits come with substantial trade-offs. Foremost, the additional requirement of annotating training examples with text output $\textoutput$ increases the complexity of data curation. Furthermore, as previously established, \gls{CoT} reasoning can drastically increase inference latency, thereby imposing severe constraints on the practical deployment of such models in real-world applications.
\end{observation}

\subsubsection[Initial Trajectory Prediction for CoT Reasoning \Mthreesymb]{Initial Trajectory Prediction for \gls{CoT} Reasoning \Mthreesymb}
\label{subsubsec:inittraj}
Different from using only $\textoutput$ for the \gls{CoT} reasoning, few approaches leverage it to assess or refine a trajectory that the model initially predicted as $\InitialPrediction$. Such methods can also leverage $\textoutput$ to improve $\InitialPrediction$ or $\Prediction$ further. As an earlier approach, \textbf{Agent-driver}~\citep{agentdriver} essentially extends GPT-Driver~\citep{gptdriver} (\cref{subsubsec:textoutputcot}) with a multi-step \gls{CoT} reasoning involving---notable objects (with their potential effects) and a meta-action $(\textoutput)$, and an initial trajectory $\InitialPrediction$. A predicted occupancy map from the perception module is also used to check whether \textit{ $\InitialPrediction$ resulted in a collision}. If so, $\InitialPrediction$ is corrected as $\Prediction$ , otherwise it is accepted as the final trajectory $\Prediction$. Following GPT-Driver, it is built on GPT 3.5 and employs a text interface as the vision adapter, hence it is not \gls{E2E}-trainable.
Alternatively, \textbf{FeD}~\citep{feedbackguided} first predicts a trajectory $\InitialPrediction \sim p(\PredictionRandvar|\Observation)$ with the initial system prompt $\textit{``Predict ten future locations in 2.5 seconds''}$. This is then followed by refining $\InitialPrediction$ using \textit{the feedback of the same model as the \gls{CoT} reasoning} including potential collisions, traffic light violations or deviations from the route or expert behaviour, as detailed in \cref{tab:trajonlyinputexamples}. Finally, the trajectory is refined, i.e., $\Prediction \sim p(\PredictionRandvar|\Observation, \textoutput, \InitialPrediction)$ by conditioning on the feedback as $\textoutput$ and the initial trajectory $\InitialPrediction$. As a result, unlike Agent-driver, relying on a frozen occupancy prediction model, \textit{the feedback on $\InitialPrediction$ is learned} in an \gls{E2E} manner during training, and provided as an output such that $\Prediction$ considers it. FeD also employs an expert model with privileged information, similar to DriveGPT4-v2, to augment the dataset with the demonstrations from the expert (refer to \cref{subsubsec:withoutcot}), but differently by using feature distillation from the expert. Alternatively, \textbf{Solve-VLM}~\citep{solve} is designed to have the final trajectory always in two-steps using \textit{Trajectory \gls{CoT}}. Specifically, a coarse trajectory, $\InitialPrediction$, is predicted among a set of predetermined trajectories, and then, $\InitialPrediction$ is refined to be more precise as $\Prediction$. Unlike FeD, the model does not explicitly make explanations, e.g., about potential collisions or traffic violations. Solve-VLM uses a custom \gls{VLM} by combining EVA-02~\citep{eva02} with a LLaVA-based \gls{LLM} via their proposed SQ-Former to incorporate 3D representations into the vision features before passing them to the \gls{LLM}. As Solve~\citep{solve} also transfers knowledge to another \gls{AD} model, we will further explore it in \cref{sec:improve} while discussing methods leveraging knowledge transfer.

\begin{observation}
{Models using Initial Trajectory Prediction for \gls{CoT} \Mthree}
These approaches offer the advantage of iterative trajectory refinement, thereby potentially enhancing driving performance.
By utilising exclusively the initial trajectory prediction for \gls{CoT} reasoning (i.e., without incorporating additional text-based \gls{CoT}), these methods can be adopted with computational efficiency, particularly when trajectories are generated through a single forward pass. Given that these models perform self-refinement of their outputs, the data curation does not require an additional overhead; however, the model design must incorporate appropriate mechanisms to facilitate this refinement process. A notable drawback, though, is that without text generation capabilities, these models lack direct explainability and cannot interact with the user.
\end{observation}

\begin{table*}[t]
    \setlength{\tabcolsep}{0.2em}
    \centering
    \caption{
    Characteristics of the models providing language interaction capability. If a model is trained in multiple settings such as DriveLM-Agent or Emma, then we include only one of them for clarity.
    \label{tab:modelswithlanguageinteraction}
    }
    \scalebox{0.5}{
\begin{tabular}{
>{\arraybackslash}m{0.18\linewidth}
>{\columncolor{gray!8}\arraybackslash}m{0.2\linewidth}
>{\arraybackslash}m{0.22\linewidth}
>{\columncolor{gray!8}\arraybackslash}m{0.25\linewidth}
>{\arraybackslash}m{0.2\linewidth} 
>{\columncolor{gray!8}\arraybackslash}m{0.2\linewidth}
>{\arraybackslash}m{0.2\linewidth}
>{\columncolor{gray!8}\arraybackslash}m{0.25\linewidth}
>{\arraybackslash}m{0.22\linewidth}}
    \toprule
    \centering{\textbf{Model}}&\centering{\textbf{Training Dataset}}&\centering{\textbf{Training Dataset Size}}&\centering{\textbf{Overview of Language Annotations}}&\centering{\textbf{Main Annotator for Language}}&\centering{\textbf{Augmentation of Q\&As with \gls{FM}}}&\centering{\textbf{An Example Question}}&\centering{\textbf{Model Training}}&\textbf{Language Evaluation}\\ 
    \midrule
    \textit{DriveGPT4}\par{}\citep{drivegpt4}&BDD-X +Custom Q\&As +Additional non-\gls{AD} datasets&16K BDD-X Q\&As are enhanced with 40K Chat-GPT Q\&As as the custom dataset&Action description, action justification, scene description&Human for BDD-X, and Chat-GPT for the custom&Questions of BDD-X &Is there any risk to the ego vehicle?&Step 1. Vision-language alignment by only training vision adapter\newline
    Step 2. Mix-fine-tuning by combining both \gls{AD} and non-\gls{AD} datasets&CIDEr, BLEU, ROUGE, GPT-Score \\ \midrule
    \textit{DriveLM-Agent}\par{}\citep{drivelm}&DriveLM-nuScenes&4K front view images with 300K Q\&A pairs&Video clip level scene descriptions. \newline Image-level perception, prediction, planning, behaviour and motion \glspl{VQA}&\centering{Humans}&\centering{Questions}&What object should the ego vehicle notice first / second / third when the ego vehicle is getting to the next possible location?&Fine-tuning using the training set&SPICE and GPT-Score for perception, planning and prediction \glspl{VQA}, and classification accuracy for behaviour \gls{VQA}. \\ \midrule
    \textit{Emma}\par{}\citep{emma}&Custom Dataset +Waymo Open Motion Dataset &Custom Dataset: 203K hours of driving \newline 
    Waymo Open Motion Dataset: 572 hours of driving
    &3D object detection, drivable road graph estimation, road blockage estimation & 
    Using perception annotations labeled by humans&\centering{\xmark}&Is the road ahead temporarily blocked?&Step 1. Pretraining on the large-scale custom dataset \newline
    Step 2. Fine-tuning using Waymo Open Motion Dataset & Task metrics such as Precision-Recall curve for object detection\\ \midrule
    \textit{OpenDriveVLA}\par{}\citep{OpenDriveVLA}&nuCaption +nuScenes-QA +nu-X&nuScenes dataset with $\sim$4 hours of driving scenarios in 700 video clips&nuCaption: Description of the scene, objects and potential risks \newline
    nuScenes-QA: Scene understanding Q\&As including existence, counting, query-object, query-status and comparison-type questions  \newline
    nu-X: Action description and justification&nuCaption: LLaMA-Adapter and GPT-4
    \newline
    nuScenes-QA: Perception annotations by humans 
    \newline
    nu-X: Human&nuCaption: Both questions and answers \newline 
    nuScenes-QA: None
    \newline
    nu-X: Only answers&Are there any cars to the front right of the stopped bus?&Step 1. Vision-language alignment \newline
    Step 2. Fine-tuning for \glspl{VQA} \newline Step 3. Fine-tuning for motion prediction \newline Step 4. Fine-tuning for trajectory planning&nuCaption: BLEU and BERT-Score \newline nuScenes-QA: Accuracy for each question type \newline nu-X: CIDEr, BLEU, METEOR and ROUGE \\ \midrule
    \textit{DiMA-MLLM}\par{}\citep{dima}&DriveLM-nuScenes extended with the remaining images in nuScenes&nuScenes dataset with $\sim$4 hours of driving scenarios in 700 video clips&Video-clip level scene descriptions. Image-level perception, prediction, planning, behaviour and motion \glspl{VQA}&Human, and Llama-3-70B for the extension&No additional augmentation for the extension&What are the future movements of the agents to the back right of the ego car?&Step 1. Training only the E2E model \newline Step 2. Fine-tuning E2E model and \gls{VLM} jointly&Qualitative evaluation \\  \midrule
    \textit{Omni-Q/L}\par{}\citep{omnidrive}&
    OmniDrive&nuScenes dataset with 700 video clips for training ($\sim$4 hours of driving scenarios)&Scene descriptions, attention, counterfactual reasoning (as in the example question), decision making and planning, general conversation&\centering{GPT-4}&\centering{\xmark}&If I decide to accelerate and make a left turn, what could be the consequences?&Step 1. Vision-language alignment using 2D perception tasks \newline Step 2. Fine-tuning the model& CIDEr for language evaluation, Average Precision and Average Recall for counterfactual reasoning \\ \midrule
    \textit{Orion}\par{}\citep{ORION}&Chat-B2D&B2D-Base dataset with $\sim$7 hours of driving scenarios and 2.11M Q\&As in  950 video clips&Scene description, behaviour description of critical objects, meta-actions and action reasoning of the ego, recall of essential historical information&Based on CARLA simulator state and Qwen2VL-72B&\centering{\xmark}&How has the current speed changed compared to the previous frames?&Step 1. Vision-language alignment using 2D perception tasks \newline Step 2. Language-action alignment without \gls{VQA} dataset \newline Step 3. Fine-tuning the model with \gls{VQA} dataset &Qualitative evaluation\\
    \midrule
    \textit{SimLingo}\par{}\citep{SimLingo}&Custom&3.1M front images are annotated in DriveLM style &Image-level perception, prediction, planning, behaviour and motion \glspl{VQA}&Based on CARLA simulator state&Questions and Answers&Is there a traffic light in the scene?&Fine-tuning using the training set& SPICE and GPT score on DriveLM dataset\\ 
    \bottomrule
    \end{tabular}
     }
\end{table*}

\subsection{Models Providing Additional Capabilities}
\label{subsec:additionalcapabilities}
We now discuss models that provide language or action capabilities in addition to trajectory planning.\footnote{Since SimLingo is the only approach providing both capabilities \Msixsymb \label{langactinteraction}, we present its language  capabilities in \cref{tab:modelswithlanguageinteraction} and discuss SimLingo in detail in \cref{subsec:vlas}, instead of reserving a section for it.}

\subsubsection{Models Providing Language Interaction Capability \Mfoursymb}
\label{subsec:langinteraction}
Clearly, the ability to interact via language is a remarkable feature enabled by the language interface of \glspl{LLM}. 
\cref{tab:modelswithlanguageinteraction} provides an overview of approaches using language interaction, including characteristics regarding their training datasets, training approaches, and language evaluations. Below, we summarise our observations regarding their training datasets and evaluations. 
\textbf{Language Interaction Capability Datasets.} Models with language interaction capability generally require pairs of \glspl{VQA} for training, which are not part of standard driving datasets such as nuScenes~\citep{nuscenes}. Hence, after creating the necessary question templates, there are three main approaches to annotate these datasets, depending on whether the dataset is real-world or simulated:
\begin{compactitem}
    \item For simulation datasets, usually curated using CARLA~\citep{carla}, predefined answer templates are populated by the \textit{simulation state} including agents' velocities and positions, the weather, junctions, and traffic lights. This approach is used in DriveLM-CARLA~\citep{carla} and the SimLingo dataset~\citep{SimLingo}, and is a relatively easy approach resulting in accurate language annotations.
    \item For real-world datasets, the desired details of the scene are usually not available. As a result, some approaches rely on \textit{human annotations} such as DriveLM-nuScenes~\citep{drivelm}. Though this results in accurate annotations, it requires significant resources to annotate a large dataset.
    \item Alternatively, \glspl{FM} such as GPT-4~\citep{gpt4} are frequently used to create language labels. Specifically, the question templates are provided to the \gls{FM} along with a system prompt and sensor data to obtain the answers. This approach is used to annotate both simulation datasets~\citep{ORION} and real-world datasets~\citep{dima}. As this is an \textit{automated approach}, it is efficient, though manual inspection might be necessary to ensure the quality of the annotations.
\end{compactitem} 
After constructing the \gls{VQA} dataset, several methods \textit{augment} questions or answers using an \gls{LLM} to increase language variability. Furthermore, the size of the training dataset varies significantly across methods. For instance, while DriveGPT4 is trained on 56K \glspl{VQA}, Orion uses 2.11M \glspl{VQA} for training. Additionally, the \glspl{VQA} generally focus on perception, prediction and planning as subtasks that allow a model to achieve good driving performance. 
DriveLM~\citep{drivelm} and OmniDrive~\citep{omnidrive} are proposed as \gls{VQA} benchmarks. The DriveLM dataset includes \gls{VQA} pairs for these tasks, such as \textit{``what object should the ego vehicle notice first when the ego vehicle is getting to the next possible location?''}. Alternatively, OmniDrive stands out with \textit{counterfactual \glspl{VQA}}, such as \textit{``If I decide to accelerate and make a left turn, what could be the consequences?''}, where the counterfactual questions are designed using templates relying on the meta-actions such as \textit{accelerate and make a left turn} in this example. Consequently, the answers are obtained by a combination of a rule-based checklist and GPT-4.

\textbf{Evaluating Language Interaction Capability.} For evaluating the quality of the generated text, the methods commonly employ measures from natural language processing literature, comparing machine-generated text with reference text. These measures include BLEU~\citep{bleu}, ROUGE~\citep{rouge}, METEOR~\citep{meteor} as well as CIDEr~\citep{cider} and SPICE~\citep{spice}, which are specifically introduced for image captioning. Furthermore, model-based evaluation measures are also utilised. For example, GPT-Score~\citep{gptscore}, used by \cite{drivelm}, is obtained by prompting Chat-GPT with the question, the reference answer, and the machine-generated answer, and asking for a numerical score about the accuracy of the machine-generated answer. Alternatively, \cite{OpenDriveVLA} use BERT-Score~\citep{bertscore} to compare the similarity of the machine-generated text with the reference in the BERT embedding space~\citep{bert}.

\textbf{A Discussion of Existing Approaches.} Here, we highlight the key aspects of the trajectory planning approaches providing language capability. Please refer to \cref{tab:trajectoryplanningmethods} and \cref{tab:modelswithlanguageinteraction} for the details on their model design and language interaction capability, respectively. \textbf{DriveGPT4}~\citep{drivegpt4} aims to retain the capabilities of the \gls{LLM} while training a trajectory planning model. To achieve this, they found it useful to \textit{keep non-\gls{AD} related \glspl{VQA}} in addition to 56K driving-related \glspl{VQA} during training. As the model is based on a custom \gls{VLM} obtained by combining a video encoder~\citep{valley} with Llama2~\citep{llama2}, the training of the model is handled in two stages, vision-language alignment and model fine-tuning, following \cref{fig:designdecisions}(d)(Right). Alternatively, \textbf{DriveLM-Agent}~\citep{drivelm}, the proposed baseline of the DriveLM benchmark, fine-tunes BLIP-2~\citep{blip2}, a \gls{VLM} based on the Flan-T5 \gls{LLM}~\citep{chung2024scaling}, on the \glspl{VQA} of the benchmark in a single stage. \textbf{Emma}~\citep{emma}, based on Gemini \gls{VLM}~\citep{gemini}, uses a large dataset of 203K hours of driving scenes for the \textit{pretraining} of the model. This pretrained model is then fine-tuned on specific domains for adaptation, such as on NuScenes dataset~\citep{nuscenes}. Instead of the conversation type questions and answers, the language capabilities of Emma seems to be limited to the questions regarding a predetermined set of perception tasks including object detection, road graph estimation, and road blockage detection. Different from the existing approaches, \textbf{OpenDriveVLA}~\citep{OpenDriveVLA} and \textbf{DiMA-MLLM}~\citep{dima} rely on BevFormer~\citep{bevformer} as the vision encoder to \textit{effectively extract a 3D representation of the scene} to address the limitation of the \gls{FM}-based vision encoder such as CLIP~\citep{clip,siglip} or EVA~\citep{eva02}. However, this can also cause a potential disadvantage in comparison to such models, considering that BevFormer is not pretrained on a large dataset. \textbf{Omni-Q} and \textbf{Omni-L}~\citep{omnidrive} share the same architecture, where EVA-02-L~\citep{eva02} vision encoder is combined with LLaVA-v1.5 \gls{LLM}~\citep{LLaVA15}, \textit{except their vision adapters}. Specifically, Omni-L relies on a linear layer following LLaVA~\citep{LLaVA} (see \cref{fig:designdecisions}(b)(Middle)) while Omni-Q employs a Q-Former supervised by 3D perception tasks (refer to \cref{fig:designdecisions}(b)(Right)). Their analyses suggest that using a \textit{linear layer is more beneficial than a Q-Former} in terms of both language capabilities and open-loop driving performance. Finally, \textbf{Orion}~\citep{ORION} combines EVA-02-L with Vicuna-1.5~\citep{vicuna15} using a Q-Former variant. The proposed Q-Former, coined as QT-Former, also considers the temporal aspect of the observations through a memory bank to improve the driving performance. Furthermore, Orion presents notable analyses on using different architectures as planning heads (refer to \cref{fig:designdecisions}(c)(Right)). Their analyses conclude that \textit{the planning head based on a variational autoencoder} perform better than using an \gls{MLP} or a diffusion model.

\begin{observation}
{Models Providing Language Interaction Capability \Mfour}
This group of approaches offers the distinct advantage of answering user questions, primarily serving to enhance model explainability and provide reassurance regarding the system's decision-making processes (i.e., trajectory). Nevertheless, similar to text-based \gls{CoT} reasoning, the generation of text output through an \gls{AR} mechanism can substantially increase inference time, thereby constraining the deployment of this functionality. Furthermore, dataset curation for these approaches presents greater complexity compared to  methods only using text-based \gls{CoT}. This is because developing language interaction capabilities typically requires each observation annotated with multiple question-answer pairs, in contrast to the singular text output per training example requisite for text-based \gls{CoT} reasoning.
\end{observation}

\begin{table*}[t]
    \setlength{\tabcolsep}{0.2em}
    \centering
    \caption{Characteristics of the models providing action interaction capability.
    \label{tab:modelswithactioninteraction}}
    \scalebox{0.59}{
\begin{tabular}{
>{\arraybackslash}m{0.18\linewidth}
>{\columncolor{gray!8}\arraybackslash}m{0.25\linewidth}
>{\arraybackslash}m{0.25\linewidth}
>{\columncolor{gray!8}\arraybackslash}m{0.30\linewidth}
>{\arraybackslash}m{0.2\linewidth} 
>{\columncolor{gray!8}\arraybackslash}m{0.2\linewidth}
>{\arraybackslash}m{0.25\linewidth}} 
    \toprule
    \centering{\textbf{Model}}&\centering{\textbf{Training Dataset (with sensors and size)}}&\centering{\textbf{Example Notice Instructions}}&\centering{\textbf{Example Action Instructions}}&\centering{\textbf{Mechanism to Avoid Misleading Instructions}}&\centering{\textbf{Model Training}}&\textbf{Instruction-following Evaluation Measures}\\ 
    \midrule
    \textit{DriveMLM}\par{}\citep{drivemlm} &280 hours of driving scenarios in CARLA. Sensors: 4 cameras (front, rear, left and right) and a lidar sensor.&\centering{\xmark}&\begin{itemize}[leftmargin=*]
        \item I’m running short on time. Is it possible for you to utilise the emergency lane to bypass the vehicles ahead?
        \item Great view on the left. Can you change to the left lane?
        \item There are obstacles ahead. Can you switch to a different lane to bypass? 
    \end{itemize}&\centering{\xmark}&Fine-tune the \gls{VLM} using the training dataset&Qualitative evaluation\\ 
    \midrule
    \textit{LMDrive}\par{}\citep{lmdrive}&3M driving scenarios collected in CARLA at 10fps ($\sim$83 hours of driving data). Sensors: 4 cameras (front, rear, left and right) and a lidar sensor&\begin{itemize}[leftmargin=*]
    \item Please watch out for the pedestrians up ahead.
\item Be mindful of the vehicle crossing on a red light to your left.
\item Please be alert of the uneven road surface in the vicinity ahead.
\end{itemize} 
&\begin{itemize}[leftmargin=*]
\item Feel free to start driving.
\item Find your way out at the first exit on the roundabout, please. 
\item At the forthcoming T-intersection, execute a right turn. Just head for the left lane. Maintain your course along this route.
\end{itemize} &\centering{\cmark}&Step 1. Train vision encoder with perception tasks using the front image \newline
Step 2. The model is trained end-to-end&Driving performance is estimated while the model is provided action and notice instructions. LangAuto benchmark is proposed for this purpose in the same paper.\\ 
    \midrule
    \textit{SimLingo}\par{}\citep{SimLingo} &3.1M front images ($\sim$215 hours of driving data) collected in CARLA at 4fps&\centering{\xmark}&\begin{itemize}[leftmargin=*]
        \item Gently press the brakes.
        \item Hit the vehicle Ford Crown.
        \item Direct one lane to the left.
    \end{itemize} & \centering{\cmark}
    &Fine-tune the \gls{VLM} using the training dataset &Accuracy of the model for each type of action interaction \\ 
    \bottomrule
    \end{tabular}
     }
\end{table*}

\subsubsection{Models Providing Action Interaction Capability \Mfivesymb}
\label{subsec:vlas}
We now review models that can plan a trajectory by considering instruction from the user. We present an overview of these methods in terms of training dataset, training approach and evaluation in \cref{tab:modelswithactioninteraction}.

\textbf{Action Interaction Capability Datasets.} All three approaches in this category (refer to \cref{fig:Taxonomy}) are trained and tested on \textit{synthetic datasets} collected using CARLA simulator~\citep{carla}. This is likely because annotating data with pairs of instructions and actions is relatively easier for synthetic data as the simulator state includes comprehensive information about the scene. Furthermore, unlike other approaches discussed in this paper, both DriveMLM and LMDrive rely on both lidar and camera. Additionally, LMDrive stands out with the functionality to consider \textit{notice instructions} such as \textit{watch the tunnel coming up}. \cite{lmdrive} introduce a specific benchmark called \textbf{LangAuto-Notice} to measure the performance of the model to respond to such notice instructions, demonstrating that LMDrive effectively leverages notice information to improve driving performance. Another crucial capability that such models are expected to have is a mechanism to avoid \textit{misleading instructions}, as such instructions can result in unsafe consequences. Ideally, upon capturing a misleading instruction, the model should reject it and follow a safe trajectory. Both LMDrive and SimLingo incorporate this crucial capability into their models.

\textbf{Evaluating Action Interaction Capability.} Unlike language interaction, the evaluation measures for action interaction are \textit{not yet well-established}. In addition to the qualitative evaluation, which is limited in terms of the number of examples and tends to have a selection bias, two different approaches are adopted to evaluate action interaction. \cite{lmdrive} design a benchmark in which the model is instructed solely by the user, similar to an extended version of the navigational commands, where the standard CARLA performance measures are reported. 
As an alternative, \cite{SimLingo} directly measure the \textit{percentage of the trajectories that is following the given instruction}, namely, the accuracy of the model to follow instructions.

\textbf{A Discussion of Existing Approaches.} 
\cref{tab:modelswithactioninteraction} provides an overview of the approaches with action interaction capability.
\textbf{DriveMLM}~\citep{drivemlm} combines EVA~\citep{EVA} and GD-MAE~\citep{gdmae} as vision and lidar encoders, respectively, with Llama \gls{LLM}~\citep{llama} through Q-Former modules. The model is then supervised to output four predefined commands (i.e., \{keep, accelerate, decelerate, stop\}) for controlling the accelerator, and five steering actions (i.e, \{follow, left change, right change, left borrow, right borrow\}). Similarly, \textbf{LMDrive}~\citep{lmdrive} processes inputs from multiple cameras and lidar using a multimodal vision encoder including ResNet~\citep{ResNet} and PointPillars~\citep{pointpillars}. This encoder is initially pretrained using object detection, traffic light status classification, and trajectory planning before it is combined with Llama~\citep{llama} where the model is supervised for trajectory planning. Unlike other approaches, the \gls{LLM} is kept frozen during the training. To determine if \textit{the given user instruction is completed}, the model predicts an additional flag, making LMDrive the only approach with this functionality. Finally, \textbf{SimLingo}~\citep{SimLingo} is built on Mini-InternVL \gls{VLM}~\citep{miniinternvl} as an extension to CarLLaVA (\cref{subsec:trajplanningonly}). Different from other approaches, SimLingo has \Msixsymb \textit{both language and action interaction capabilities}, as well as the option to use \gls{CoT} reasoning (please refer to our discussion on inference time-accuracy trade-off of \gls{CoT} reasoning in \cref{subsubsec:textoutputcot} for further details). The model is trained in a single stage on a dataset including training examples for these functionalities. The instruction following capability, coined as \textit{action dreaming}, aims to align the predicted trajectory with the natural language instructions. Consequently, it is shown that action dreaming helps improving the driving performance more than the pure language tasks, i.e., \gls{VQA} and \gls{CoT} reasoning, as shown in the paper. SimLingo also pay attention to the resampling of the dataset via carefully creating data buckets for predefined driving characteristics and then assign different sampling ratios to each bucket. 

\begin{observation}
{Models Providing Action Interaction Capability \Mfive}
The approaches with only action interaction capability offer the distinct advantage of trajectory planning conditioned on user instructions, serving as a potentially valuable additional feature. Since this functionality itself does not require generating text output, such models can be deployed efficiently in real-world systems. However, the dataset curation protocols for developing these models remain insufficiently investigated for real-world systems. Essentially, the model requires training with a substantial volume of text input (user instructions) paired with corresponding output trajectories. Crucially, the text input should exhibit sufficient variability to enable robust responses to diverse user instructions. Akin to other subgroups lacking text generation, these models could inherently suffer from a deficiency in explainability especially when \gls{CoT} reasoning is not employed either.
\end{observation}
\section{Foundation Models Guiding Trajectory Planning}
\label{sec:improve}
\begin{figure*}[t]
        \centering
        \includegraphics[width=\textwidth]{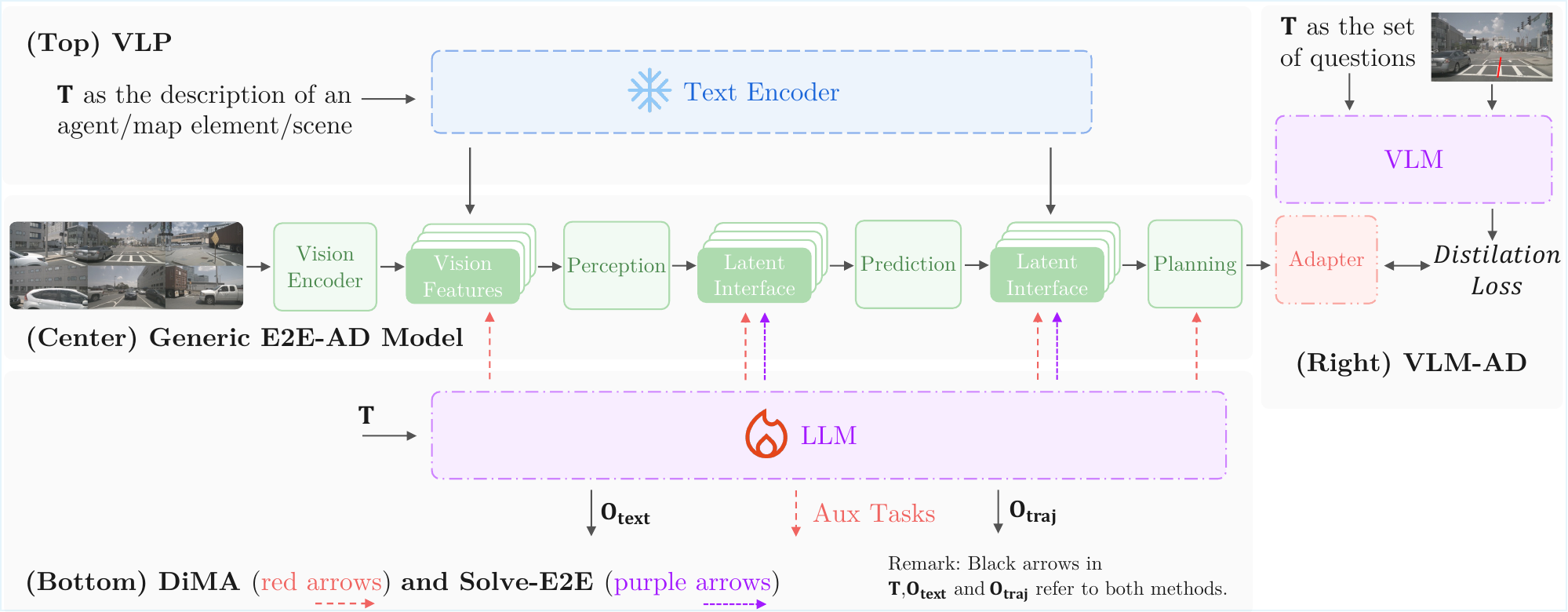}
        \caption{Overview of the approaches that uses an \gls{FM} only during training for knowledge distillation. \textbf{(Center)} A generic \gls{E2E}-\gls{AD} model (please refer to \cref{sec:background} for the details.) \textbf{(Top)} VLP prompts the model with the agent/map elements/scene descriptions to align the representations of CLIP text encoder with those of the \gls{E2E}-\gls{AD} model. The arrows from the text encoder represent the representations. \textbf{(Right)} VLM-AD prompts a \gls{VLM} with the privileged information (the red line on the input image) and a few questions to align the representations of the \gls{E2E}-\gls{AD} model. \textbf{(Bottom)} DiMA and Solve jointly trains \gls{E2E}-\gls{AD} model and the \gls{FM}, where the \gls{LLM} makes predictions by conditioning on the representations of the \gls{E2E}-\gls{AD} model.  Please refer to \cref{subsec:onlytraining} for further details.
        }
        \label{fig:distillationmethods}
\end{figure*}

An alternative way of utilising \glspl{FM} for trajectory planning is to use an existing approach for \gls{AD}, such as a modular or an \gls{E2E} one (see \cref{fig:adtypes}(a,b)), and transfer the knowledge of a chosen \gls{FM} into it either only during training or during both training and inference.
A key distinction among these approaches come from their choice of using \gls{FM} during inference which, of course, results in increased computational cost and memory requirements. 
Considering that, based on our taxonomy (refer to \cref{fig:Taxonomy}), we discuss the models using knowledge distillation only during training in \cref{subsec:onlytraining}, and those requiring an \gls{FM} also for inference in \cref{subsec:alsoinference}. 

\subsection{Models Using Knowledge Distillation Only During Training \Msevensymb}
\label{subsec:onlytraining}
As this group of approaches carries out knowledge distillation only during training, they do not need the \gls{FM} for inference. \cref{fig:distillationmethods} provides an overview of the existing approaches within this category. Among the first approaches, \textbf{VLP}~\citep{VLP} aims to align the representations of an \gls{E2E}-\gls{AD} model with an off-the-shelf CLIP~\citep{clip}. To obtain CLIP representations, only the text encoder of the CLIP is used. Specifically, the text encoder is prompted by (i) the descriptions of each agent and map elements such as their semantic class and location, and (ii) a planning prompt, such as \textit{``the self-driving car is driving in an urban area. It is going straight. Its planned trajectory is $\Prediction$''}. Then, these representations are distilled into the \gls{E2E}-\gls{AD} model in two levels as shown in \cref{fig:distillationmethods}(Top), using contrastive learning. While the agent and map element representations are distilled into vision features of the \gls{E2E}-\gls{AD} model, the representation of the planning prompt is incorporated into its ego query as the input to the planning module.
Similar to VLP, \textbf{VLM-AD}~\citep{VLMAD} keeps the \gls{FM} frozen but instead prompts a \gls{VLM}, GPT-4o~\citep{gpt4}, with (i) the front image including the privileged information of the future trajectory of the ego shown by the red line on the image in \cref{fig:distillationmethods}(Right), and (ii) six different questions to obtain a meta-action and its reasoning from the \gls{VLM}. These responses are then structured using one-hot encoding and a CLIP text encoder, and an adapter layer is appended to the \gls{E2E}-\gls{AD} model to predict this structured output. Finally, cross-entropy loss is used for knowledge distillation.

Different from the previous approaches, \textbf{DiMA}~\citep{dima} and \textbf{Solve}~\citep{solve} train \gls{E2E}-\gls{AD} and \gls{LLM} jointly for knowledge distillation. Specifically, as shown in \cref{fig:distillationmethods}(Bottom), both approaches propagate representations of the \gls{E2E}-\gls{AD} model to the \gls{LLM} of LLaVA-v1.5~\citep{LLaVA15}, such that the \gls{LLM} predicts the trajectory and a text output by conditioning on these representations. While Solve propagates only the representations after perception, DiMA aims for a more comprehensive distillation. As a result, in DiMA, (i) \gls{BEV} features, agent/map representations, and the ego query are all propagated to the \gls{LLM} (red arrows from \gls{LLM} to the blue interfaces in \cref{fig:distillationmethods}(Bottom)), and (ii) a distillation loss is introduced to align the representations of the \gls{LLM} and the planning module of the \gls{E2E}-\gls{AD} (red arrow from the \gls{LLM} to the planning module in \cref{fig:distillationmethods}(Bottom)). Furthermore, DiMA includes auxiliary tasks such as masked \gls{BEV} token reconstruction for effective representation learning during training. Both approaches fine-tune the \gls{LLM} using \gls{LoRA}~\citep{lora} for the trajectory planning task.

\begin{observation}
{Models Using Knowledge Distillation Only During Training \Mseven}
These approaches are unique in this review by their elimination of \gls{FM} dependency during inference. Consequently, the resulting \gls{AD} models typically have fewer parameters and exhibit a higher inference rate, thereby facilitating the deployment of computationally efficient systems. However, this independence from \glspl{FM} excludes access to natural language interaction capabilities. Specifically, such models cannot process user questions or instructions, and are constrained in terms of model explainability. Moreover, the complexity of the dataset curation process varies depending on whether the \gls{FM} is jointly trained and the specific type of knowledge being distilled, which often necessitates additional input-output pair annotations.
\end{observation}

\begin{table*}[t]
\setlength{\tabcolsep}{0.4em}
\renewcommand{\arraystretch}{1.4}
\small
\centering
\caption{Design choices of the models that transfer knowledge from an \gls{FM} during inference.
\label{tab:knowledgetransferduringinference}
\includegraphics[height=0.015\textheight]{Images/Frozen.png}: Frozen, 
\includegraphics[height=0.015\textheight]{Images/Finetune.png}: Standard fine-tuning, 
\includegraphics[height=0.015\textheight]{Images/LoRA.png}: LoRA fine-tuning. 
Broadly speaking, the last three columns correspond to the feature volumes (latent interfaces in blue) in \cref{fig:adtypes}(b). 
For \textit{FasionAD++}, the used \gls{FM} type for the main experiments is not clear in the paper, hence it is marked as not available (N/A).}
\scalebox{0.60}{
\begin{tabular}{
>{\centering\arraybackslash\hspace{0pt}}m{0.02\linewidth}
>{\hspace{0pt}}
>{\arraybackslash}m{0.17\linewidth}
>{\arraybackslash}m{0.18\linewidth}
>{\arraybackslash}m{0.23\linewidth}
>{\arraybackslash}m{0.34\linewidth}
>{\arraybackslash}m{0.26\linewidth} 
>{\centering\arraybackslash}m{0.1\linewidth}
>{\centering\arraybackslash}m{0.1\linewidth}
>{\centering\arraybackslash}m{0.1\linewidth}}
\toprule
&\centering{\textbf{Model}} &
\centering{\textbf{Sensor Input to \gls{FM} ($\FMObservation$)}} &
\centering{\textbf{Used \gls{FM}}} &
\centering{\textbf{Transferred Knowledge from \gls{FM} at Inference ($\LatentRepresentation$)}} &
\centering{\textbf{How to Encode $\LatentRepresentation$}} &
\multicolumn{3}{c}{\textbf{$\LatentRepresentation$ Transferred Into}} \\
\cmidrule{7-9}
&&&&&&\textbf{Vision Features} & \textbf{Prediction Input} & \textbf{Planning Input} \\
\midrule
1&\textit{VLM-E2E}\par{}\citep{vlme2e} & Front camera & BLIP-2\includegraphics[height=0.02\textheight]{Images/Frozen.png} & Scene description & CLIP text encoder + \gls{MLP} & \cmark & & \\
2&\textit{DME-Driver}\par{}\citep{DME-Driver} & Front camera & LLaVA\includegraphics[height=0.02\textheight]{Images/Finetune.png} & Scene description, driver’s gaze, and driver’s logic & BERT & & \cmark & \cmark \\
\midrule
3&\textit{Senna-E2E}\par{}\citep{Senna} & 360$^{\circ}$ cameras & ViT-L + Vicuna-v1.5\includegraphics[height=0.02\textheight]{Images/Finetune.png} & Meta-action & Learnable embedding layer & & & \cmark \\
4&\textit{DiffVLA}\par{}\citep{diffvla} & 360$^{\circ}$ cameras & ViT-L + Vicuna-v1.5\includegraphics[height=0.02\textheight]{Images/Finetune.png} & Meta-action & One-hot encoding & & & \cmark \\
5&\textit{DriveVLM-Dual}\par{}\citep{drivevlm}& Front camera & Qwen-VL\includegraphics[height=0.02\textheight]{Images/Finetune.png} & Trajectory & None & & & \cmark \\
6&\textit{Solve-E2E-Async}\par{}\citep{solve} & 360$^{\circ}$ camera features & LLaVA-1.5-LLM\includegraphics[height=0.02\textheight]{Images/LoRA.png} & Trajectory & None & & & \cmark \\
7&\textit{DiMA-Dual}\par{}\citep{dima} & 360$^{\circ}$ camera features & LLaVA-1.5-LLM\includegraphics[height=0.02\textheight]{Images/LoRA.png} & \gls{LLM} planning features in the last layer & None & & & \cmark \\
8&\textit{HE-Drive}\par{}\cite{hedrive}& 360$^{\circ}$ cameras & Llama3\includegraphics[height=0.02\textheight]{Images/Finetune.png} & Weights of a scoring function to select the most suitable trajectory among multiple ones & None & & & \cmark \\
\midrule
9&\textit{VDT-Auto}\par{}\citep{vdtauto}& Front camera & Qwen2-VL\includegraphics[height=0.02\textheight]{Images/Finetune.png} & \gls{VLM} features of the detected objects, meta-action, trajectory proposals & None & & & \cmark \\
10&\textit{FasionAD++}\par{}\citep{fasionad}& Front camera & N/A & Existence of predetermined objects (e.g., traffic lights, intersections, obstacles) and meta-actions & Binary encoding for object existence, learnable embeddings for meta-actions & \cmark & \cmark & \cmark \\
11&\textit{AsyncDriver}\par{}\citep{fasionad}&Vectorised encoding of the scene& Llama2\includegraphics[height=0.02\textheight]{Images/LoRA.png} & Ego states, the occupancy of the adjacent scene, the state of the traffic light, lane change and velocity decisions
& Learnable embedding layer & & & \cmark \\
\bottomrule
\end{tabular}
}
\end{table*}

\subsection{Models Using Knowledge Transfer During Inference \Meightsymb}
\label{subsec:alsoinference}
Unlike the methods in the previous section, the approaches we discuss here are the ones employing \glspl{FM} during both training and inference for more effective knowledge transfer.
\cref{tab:knowledgetransferduringinference} provides the main characteristics of these approaches, including the type of knowledge transferred from the \gls{FM} and how this knowledge is integrated into the \gls{AD} model. In the following, we elaborate on these approaches based on the type of knowledge each transfers to the \gls{AD} model, which is usually a (i) scene description involving perception or prediction features, (ii) a planning decision such as a meta-action or trajectory, or (iii) both.

\textbf{Methods that Transfer Scene Description.} These methods aim to transfer scene descriptions, such as \textit{``a black van driving in the ego lane, away from the ego car''}, taken from \textbf{VLM-E2E}~\citep{vlme2e}. Specifically, \cite{vlme2e} prompt an off-the-shelf BLIP-2~\citep{blip2} with a front image of a driving scene to obtain such a description. As this description is essentially a text, CLIP text encoder~\citep{clip} converts it into a text representation. Then, the representation is mapped to shifting and scaling factors using \glspl{MLP} to update the \gls{BEV} features of the \gls{E2E}-\gls{AD} model, in this case UniAD~\citep{uniad}. \textbf{DME-Driver}~\citep{DME-Driver} follows a similar approach, including knowledge of the scene description as well as \textit{the driver's gaze} and the driver's logic on the taken action, such as the presence of the pedestrians for slowing down. Particularly, given the front view of the driving scene, LLaVA~\citep{LLaVA} is fine-tuned to produce the text output, which is converted into embeddings using BERT~\citep{bert}. Finally, \gls{BEV} features of the \gls{E2E}-\gls{AD} model attend to these embeddings through cross-attention before being fed into the occupancy prediction and planning modules of UniAD~\citep{uniad}. Though these methods show that such a knowledge transfer is useful, they do not guide the \gls{E2E}-\gls{AD} model \textit{explicitly} with a planning decision such as a meta-action or a trajectory, which we discuss next.

\textbf{Methods that Transfer Planning Decisions.} These methods transfer the planning decisions of the \gls{FM} to the \gls{AD} model, usually in the form of a meta-action or trajectory. For example, \textbf{Senna-E2E}~\citep{Senna} transfers the meta-action of the \gls{VLM} by fine-tuning it specifically for this task. The model is fine-tuned in multiple stages to \textit{progressively} specialise the \gls{VLM} for planning: In driving fine-tuning, the \gls{VLM} is supervised with \glspl{VQA} in driving scenarios, followed by planning fine-tuning for meta-action classification. After the \gls{VLM} is trained, it is frozen and \textit{the meta-action of the \gls{VLM} is converted into an embedding} using a learnable layer. Then, in order to benefit from the knowledge of the \gls{VLM}, the planning query in the \gls{AD} model, VADv2~\citep{vadv2}, attends to this meta-action embedding of the \gls{VLM}.   \textbf{DiffVLA}~\citep{diffvla} also relies on Senna-VLM but with two main differences. First, instead of a learnable embedding layer, a one-hot encoding of the meta-action is passed to the planning module. Second, the VADv2 planner is replaced by a \textit{diffusion planner}~\citep{DiffusionDrive}, which generates the trajectory by conditioning on this meta-action encoding as well as the \gls{BEV} features, map and object queries \textit{sequentially}. The approach is trained and evaluated using the NAVSIM v2 benchmark~\citep{cao2025pseudo}.

Differently, some approaches, including DriveVLM, Solve and DiMA, fine-tune their \glspl{VLM} for the trajectory planning task. This manifests itself as an additional advantage of combining the planned trajectories from the \gls{AD} model and the  \gls{VLM} for potentially improving the driving performance. To begin with, \textbf{DriveVLM-Dual}~\citep{drivevlm} uses the trajectory of the \gls{VLM} as \textit{the input query of the planning module} of the \gls{E2E}-\gls{AD} model, such that the planning module refines it further. This approach is incorporated into UniAD~\citep{uniad}, VAD~\citep{vad} and AD-MLP~\citep{ADMLP}. Similarly, \textbf{Solve-E2E-Async}~\citep{solve} uses the trajectory of the \gls{LLM} as an additional query of the planning module in the \gls{E2E}-\gls{AD} model, yet \textit{asynchronously}. That is, to account for the longer inference time of the \gls{VLM}, (i) the prediction horizon of the \gls{VLM} is designed to be longer than that of the \gls{E2E}-\gls{AD} model, and (ii) the \gls{E2E}-\gls{AD} model uses the last predicted trajectory of the \gls{VLM} as the additional planning query. Alternatively, \textbf{DiMA-Dual}~\citep{dima} transfers \textit{the representation yielded by \gls{VLM} for trajectory planning}. Specifically, max-pooling is used on the last layer features of the planning representations obtained in the \gls{VLM} and \gls{E2E}-\gls{AD} models, and then the \gls{E2E}-\gls{AD} model predicts the trajectory from the resulting pooled features. 
Finally, as a different approach, \textbf{HE-Drive}~\citep{hedrive} fine-tunes Llama3~\citep{llama3} to output the weights of a function scoring candidate trajectories. Specifically, \cite{hedrive} use a diffusion-based planner to sample multiple candidate trajectories, each of which is scored considering various driving-related factors such as collision risk, target speed compliance and comfort. For example, if there is a stopped vehicle in front of ego and the ego needs to slow down, the weight of the target speed compliance is increased by the \gls{VLM}, thereby helping to select the best trajectory. 

\textbf{Methods that Transfer Scene Description and Planning Decision.} Some approaches transfer both a scene description and a planning decision. \textbf{VDT-Auto}~\citep{vdtauto} fine-tunes Qwen2-VL~\citep{qwenvl} on driving-related \glspl{VQA} to output object descriptions, as well as a meta-action and trajectory. The corresponding representations in the \gls{VLM} are then transferred to the \gls{AD} model by freezing the \gls{VLM}. Specifically, these embeddings are used as inputs to a \textit{diffusion-based planner}~\citep{DiffusionDrive}, which refines the trajectories conditioned on these embeddings and the vision features. Alternatively, \textbf{FasionAD++}~\citep{fasionad} guides the \gls{VLM} to output (i) a planning state including binary variables indicating the existence of, e.g., traffic lights, obstacles, intersections; and (ii) a meta-action. The planning state updates BEV features, agent, map and ego queries through an adapter layer, while the meta-action is incorporated into the ego features only. As a result, FasionAD++ provides knowledge transfer at multiple levels into the \gls{E2E}-\gls{AD} model. Additionally, FasionAD++ takes the inefficiency of the \gls{VLM} into account by referring to the \gls{VLM} only when the \textit{uncertainty of the \gls{AD} model} increases beyond a certain threshold. Unlike other methods that we discuss in this section, \textbf{AsyncDriver}~\citep{chen2024asynchronous} transfers \gls{LLM} knowledge to the planning module of a modular approach (see \cref{fig:adtypes}(a)). Specifically, it fine-tunes an \gls{LLM} to predict useful information for \gls{AD} by appending it with an \textit{assistance alignment module}. Conditioned on the output of the \gls{LLM}, this module predicts relevant perception features such as traffic light states, occupancy of the adjacent lane, as well as features affecting the trajectory planning such as lane change and velocity decisions. The latent encoding of the \gls{LLM}, used as the input to the assistance alignment module, is propagated to the modular planner via a feature adapter. To align the \gls{LLM} and the modular planner, they are trained jointly on the nuPlan dataset~\citep{caesar2021nuplan}. During inference, the latent encoding from the LLM is passed \textit{asynchronously} to the modular planner to improve the planning decision, maintaining the previous high-level instruction features during intervals. 
This asynchronous connection between the separate real-time and LLM-based planners provides a balance between quality and inference speed for responses.

\begin{observation}
{Models Using Knowledge Transfer During Inference \Meight}
This category of approaches encompasses systems that integrate a \gls{FM} with an \gls{AD} model. The inference latency of such coupled architectures relies upon specific design parameters. 
For instance, continuous reliance on the \gls{FM} by the \gls{AD} model can substantially increase inference overhead, whereas selective and conditional \gls{FM} invocation can enhance average computational efficiency. Additionally, the inference time is influenced by whether the transferred knowledge is obtained following \gls{AR} steps within the \gls{FM}. Regarding explainability, the integrated \gls{FM} component might generate natural language explanations; however, ensuring semantic alignment between the \gls{AD} model's decision-making processes and the \gls{FM}'s textual outputs presents considerable challenges. Similar to approaches employing knowledge distillation, the complexity of dataset curation depends upon whether the \gls{FM} undergoes joint training and the specific nature of the information being transferred, necessitating the collection of additional text input-output pairs.
\end{observation}
\section[How Open Are the Dataset and Code of Existing Approaches?]{How Open Are Data and Code of the Existing Approaches?}
\label{sec:openness}

\begin{figure}[htpb]
        \captionsetup[subfigure]{}
        \centering
        \begin{subfigure}[b]{0.54\textwidth}
            \includegraphics[width=\textwidth]{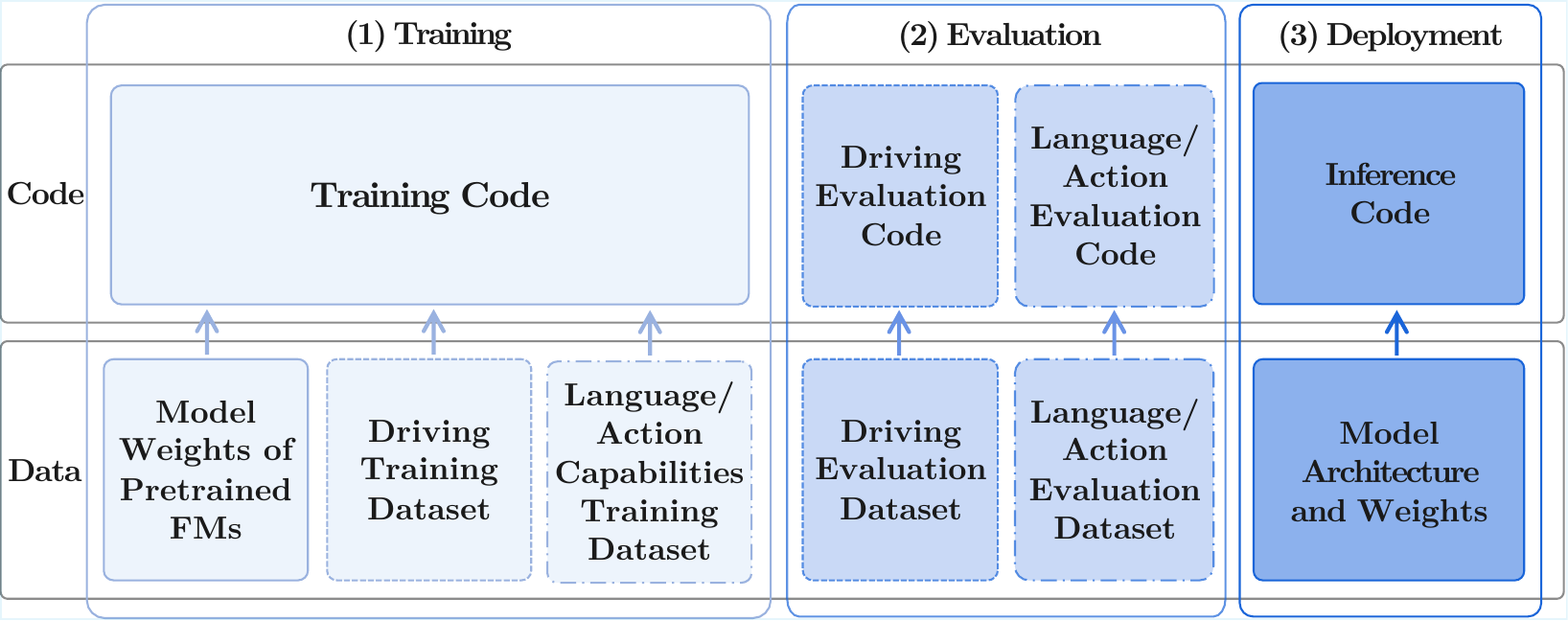}
            \caption{The code and data for each stage of development}
        \end{subfigure}
        \begin{subfigure}[b]{0.41\textwidth}
            \includegraphics[width=\textwidth]{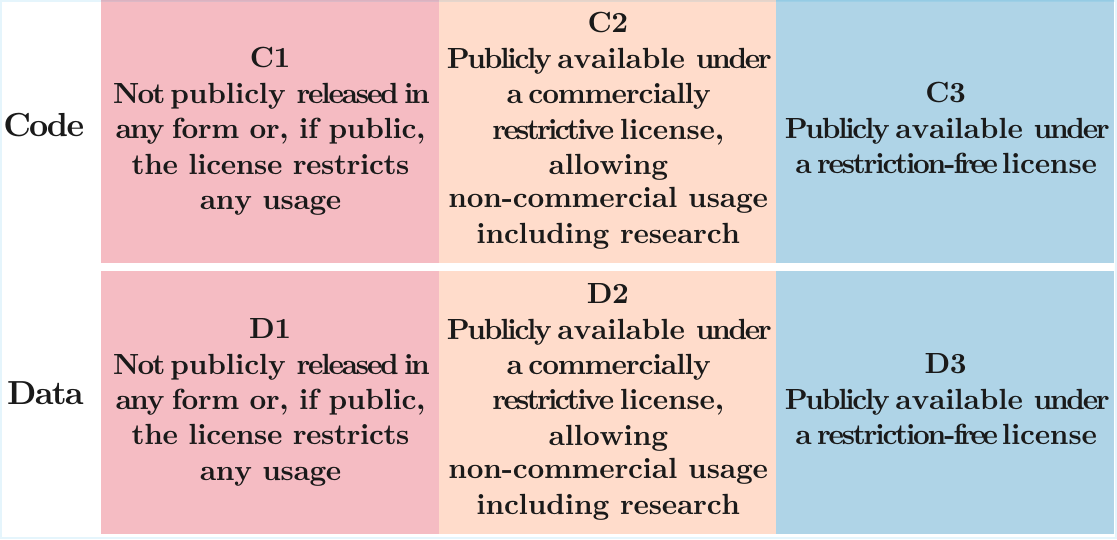}
            \caption{The levels of openness for each asset}
        \end{subfigure}
        \caption{Representation of different stages of model development, and the levels of openness.}
        \label{fig:openness_fig}
\end{figure}

\begin{table*}[!htpb]
\setlength{\tabcolsep}{0.2em}
\renewcommand{\arraystretch}{1.7}
\small
\centering
\caption{Openness characteristics.
\label{tab:openness} While every effort has been made to ensure the accuracy of this table, we cannot guarantee its completeness or correctness. The information is provided for general reference only and should not be interpreted as legal advice. If in doubt, please confirm with the corresponding authors. 
The table reflects the status as of 08 October 2025.
Entries marked as N/A indicate that the asset is ``not applicable'' to the method. We further outline the conventions we used while classifying the openness of the approaches in \cref{app:conventions}, which also clarifies the symbols * and **. We label the first method of each taxonomy group with its corresponding icon.}
\scalebox{0.46}{
\begin{tabular}{
>{\centering\arraybackslash\hspace{0pt}}m{0.02\linewidth}
>{\centering\arraybackslash\hspace{0pt}}m{0.02\linewidth}
>{\hspace{0pt}}m{0.18\linewidth}
>{\centering\hspace{0pt}}m{0.18\linewidth}
>{\centering\hspace{0pt}}m{0.18\linewidth}
>{\centering\hspace{0pt}}m{0.18\linewidth}
>{\centering\arraybackslash\hspace{0pt}}m{0.18\linewidth}
>{\centering\hspace{0pt}}m{0.18\linewidth}
>{\centering\hspace{0pt}}m{0.18\linewidth}
>{\centering\hspace{0pt}}m{0.18\linewidth}
>{\centering\arraybackslash\hspace{0pt}}m{0.18\linewidth}
>{\centering\hspace{0pt}}m{0.18\linewidth}
>{\centering\arraybackslash\hspace{0pt}}m{0.18\linewidth}}
\toprule
\multirow{3}[6]{*}{} &
\multirow{3}[6]{*}{} & 
\multirow{3}[6]{*}{\textbf{Method}} & 
\multicolumn{4}{c}{\textbf{Training}} & 
\multicolumn{4}{c}{\textbf{Evaluation}} & 
\multicolumn{2}{c}{\textbf{Deployment}} \\ 
\cmidrule(lr){4-7}
\cmidrule(lr){8-11}
\cmidrule(lr){12-13}
& & & 
\textbf{Training Code} & 
\textbf{Model Weights of Pretrained \glspl{FM}} & 
\textbf{Driving Training Dataset} & 
\textbf{Language/Action Capabilities Training Dataset} & 
\textbf{Driving Evaluation Code} & 
\textbf{Driving Evaluation Dataset} & 
\textbf{Language/Action Evaluation Code} & 
\textbf{Language/Action Evaluation Dataset} & 
\textbf{Inference Code} & 
\textbf{Model Architecture and Weights} \\ 
\Mone & 1 & \textit{CarLLaVA}\par{}\citep{CarLLaVA} & \CThreeTable\par{}{Apache2.0} & \DThreeTable\par{}LLaVA-NeXT & \DTwoTable\par{}Custom & \NATable & \CThreeTable\par{}Apache2.0 & \DThreeTable\par{}Bench2Drive & \NATable & \NATable & \CThreeTable\par{}Apache2.0 & \DOneTable \\
& 2 & \textit{DriveGPT4-v2}\par{}\citep{DriveGPT4v2} & \COneTable & \DThreeTable\par{}Qwen-0.5B & \DOneTable\par{}Custom & \NATable & \COneTable & \DThreeTable\par{}CARLA L\addition{ongest}6 & \NATable & \NATable & \COneTable & \DOneTable \\ 
& 3 & \textit{V2X-VLM}\par{}\citep{v2xvlm} & \COneTable & \DThreeTable\par{}Florence-2 & \DThreeTable\par{}DAIR-V2X & \NATable & \COneTable & \DThreeTable\par{}DAIR-V2X & \NATable & \NATable & \COneTable & \DOneTable \\ 
\Mtwo & 4 & \textit{GPT-Driver}\par{}\citep{gptdriver} & \COneTable\par{}No license & \DOneTable\par{}GPT-3.5 & \DTwoTable\par{}nuScenes & \NATable & \COneTable\par{}No license & \DTwoTable\par{}nuScenes & \NATable & \NATable & \COneTable\par{}No license & \DOneTable \\ 
& 5 & \textit{Drive-VLM}\par{}\citep{drivevlm} & \COneTable & \DThreeTable\par{}Qwen-VL & \DTwoTable\par{}nuScenes & \NATable & \COneTable & \DTwoTable\par{}nuScenes & \NATable & \NATable & \COneTable & \DOneTable \\ 
& 6 & \textit{Auto-VLA}\par{}\citep{autovla} & \COneTable & \DThreeTable\par{}Qwen2.5-VL & \DTwoTable\par{}WOMD & \NATable & \COneTable & \DThreeTable\par{}Bench2Drive & \NATable & \NATable & \COneTable & \DOneTable \\
& 7 & \textit{RAG-Driver}\par{}\citep{ragdriver} & \CThreeTable\par{}Apache2.0 & \DTwoTable\par{}Vicuna v1.5 & \DTwoTable\par{}BDD-X & \NATable & \CThreeTable\par{}Apache2.0 & \DTwoTable\par{}BDD-X & \NATable & \NATable & \CThreeTable\par{}Apache2.0 & \DOneTable \\
& 8 & \textit{S4-Driver}\par{}\citep{S4Driver} & \COneTable & \DOneTable\par{}PaLI-3 & \DTwoTable\par{}WOMD & \NATable & \COneTable & \DTwoTable\par{}nuScenes & \NATable & \NATable & \COneTable & \DOneTable \\
\Mthree & 9 & \textit{Agent-driver}\par{}\citep{agentdriver} & \CThreeTable\par{}MIT & \DOneTable\par{}GPT-3.5 & \DTwoTable\par{}nuScenes & \NATable & \CThreeTable\par{}MIT & \DTwoTable\par{}nuScenes & \NATable & \NATable & \CThreeTable\par{}MIT & \DOneTable \\
& 10 & \textit{FeD}\par{}\citep{feedbackguided} & \COneTable & \DTwoTable\par{}LLaVA-7B & \DThreeTable\par{}CARLA & \NATable & \COneTable & \DThreeTable\par{}LAV & \NATable & \NATable & \COneTable & \DOneTable \\
& 11 & \textit{Solve-VLM}\par{}\citep{solve} & \COneTable & \DTwoTable\par{}LLaVA v1.5 & \DTwoTable\par{}nuScenes & \NATable & \COneTable & \DTwoTable\par{}nuScenes & \NATable & \NATable & \COneTable & \DOneTable \\
\Mfour & 12 & \textit{DriveGPT4}\par{}\citep{drivegpt4} & \COneTable & \DTwoTable\par{}LLaMA2 & \DTwoTable\par{}BDD-X & \DOneTable\par{}Custom, no license & \COneTable\par{}No license & \DTwoTable\par{}BDD-X & \COneTable\par{}No license & \DOneTable\par{}Custom, no license & \COneTable\par{}No license & \DOneTable \\
& 13 & \textit{DriveLM-Agent}\par{}\citep{drivelm} & \COneTable & \DThreeTable\par{}BLIP-2 & \DTwoTable\par{}DriveLM & \DTwoTable\par{}DriveLM & \CThreeTable\par{}Apache2.0 & \DTwoTable\par{}DriveLM & \CThreeTable\par{}Apache2.0 & \DTwoTable\par{}DriveLM & \COneTable & \DOneTable \\
& 14 & \textit{Emma}\par{}\citep{emma} & \COneTable & \DOneTable\par{}Gemini & \DOneTable\par{}Custom & \DOneTable\par{}Custom & \COneTable & \DTwoTable\par{}nuScenes & \COneTable & \DTwoTable\par{WOD} & \COneTable & \DOneTable \\
& 15 & \textit{OpenDriveVLA}\par{}\citep{OpenDriveVLA} & \COneTable & \DThreeTable\par{}Qwen2.5 & \DTwoTable\par{}nuScenes & \DOneTable\par{}nu-X, no license & \CThreeTable\par{}Apache2.0 & \DTwoTable\par{}nuScenes & \CThreeTable\par{}Apache2.0 & \DTwoTable\par{}nuScenes-QA & \CThreeTable\par{}Apache2.0 & \DOneTable \\
& 16 & \textit{DiMA-MLLM}\par{}\citep{dima} & \COneTable & \DTwoTable\par{}LLaVA v1.5 & \DTwoTable\par{}nuScenes & \DOneTable\par{}Custom & \COneTable & \DTwoTable\par{}nuScenes & \COneTable & \DTwoTable\par{}DriveLM & \COneTable & \DOneTable \\
& 17 & \textit{Omni-Q}\par{}\citep{omnidrive} & \CTwoTable\par{}Custom & \DTwoTable\par{}LLaVA v1.5 & \DTwoTable\par{}nuScenes & \DTwoTable\par{}Custom & \CThreeTable\par{}Custom & \DTwoTable\par{}nuScenes & \CThreeTable\par{}Custom & \DTwoTable\par{}DriveLM & \CTwoTable\par{}Custom & \DThreeTable* \\
& 18 & \textit{Omni-L}\par{}\citep{omnidrive} & \CTwoTable\par{}Custom & \DTwoTable\par{}LLaVA v1.5 & \DTwoTable\par{}nuScenes & \DTwoTable\par{}Custom & \CThreeTable\par{}Custom & \DTwoTable\par{}nuScenes & \CThreeTable\par{}Custom & \DTwoTable\par{}DriveLM & \CTwoTable\par{}Custom & \DOneTable \\
& 19 & \textit{Orion}\par{}\citep{ORION} & \CThreeTable\par{}Apache2.0 & \DTwoTable\par{}{Vicuna v1.5} & \DThreeTable\par{}B2D-Base & \DThreeTable\par{}B2D-Chat & \CThreeTable\par{}Apache2.0 & \DThreeTable\par{}Bench2Drive & \CThreeTable\par{}Apache2.0 & \DThreeTable\par{}Bench2Drive & \CThreeTable\par{}Apache2.0 & \DThreeTable* \\
\Mfive & 20 & \textit{DriveMLM}\par{}\citep{drivemlm} & \COneTable & \DTwoTable\par{}LLaMA-7B & \DOneTable\par{}Custom & \DOneTable\par{}Custom & \COneTable & \DThreeTable\par{}CARLA \addition{Town0}5 & \COneTable & \DOneTable\par{}Custom & \COneTable & \DOneTable \\
& 21 & \textit{LMDrive}\par{}\citep{lmdrive} & \CThreeTable\par{}Apache2.0 & \DTwoTable\par{}{LLaVA v1.5} & \DTwoTable\par{}Custom & \DTwoTable\par{}Custom & \CThreeTable\par{}Apache2.0 & \DThreeTable\par{}CARLA \addition{Town0}5 & \CThreeTable\par{}Apache2.0 & \DTwoTable\par{}Custom & \CThreeTable\par{}Apache2.0 & \DTwoTable** \\
\Msix & 22 & \textit{SimLingo}\par{}\citep{SimLingo} & \CThreeTable\par{}{Apache2.0} & \DThreeTable\par{}{MiniInternVL} & \DTwoTable\par{}Custom & \DTwoTable\par{}Custom & \CThreeTable\par{}Apache2.0 & \DThreeTable\par{}Bench2Drive & \CThreeTable\par{}Apache2.0 & \DTwoTable\par{}DriveLM & \CThreeTable\par{}Apache2.0 & \DThreeTable \\
\Mseven & 23 & \textit{VLP}\par{}\citep{VLP} & \COneTable & \DThreeTable\par{}CLIP & \DTwoTable\par{}{nuScenes} & \NATable & \COneTable & \DTwoTable\par{}{nuScenes} & \NATable & \NATable & \COneTable & \DOneTable \\
& 24 & \textit{VLM-AD}\par{}\citep{VLMAD} & \COneTable & \DOneTable\par{}GPT-4o & \DTwoTable\par{}nuScenes & \NATable & \COneTable & \DThreeTable\par{}CARLA \addition{Town0}5 & \NATable & \NATable & \COneTable & \DOneTable \\
& 25 & \textit{DiMA}\par{}\citep{dima} & \COneTable & \DTwoTable\par{}LLaVA v1.5 & \DTwoTable\par{}nuScenes & \NATable & \COneTable & \DTwoTable\par{}nuScenes & \NATable & \NATable & \COneTable & \DOneTable \\
& 26 & \textit{Solve-E2E}\par{}\citep{solve} & \COneTable & \DTwoTable\par{}LLaVA v1.5 & \DTwoTable\par{}nuScenes & \NATable & \COneTable & \DTwoTable\par{}nuScenes & \NATable & \NATable & \COneTable & \DOneTable \\
\Meight & 27 & \textit{VLM-E2E}\par{}\citep{vlme2e} & \COneTable & \DThreeTable\par{}BLIP-2 & \DTwoTable\par{}nuScenes & \NATable & \COneTable & \DThreeTable\par{}CARLA \addition{Town0}5 & \NATable & \NATable & \COneTable & \DOneTable \\
& 28 & \textit{DME-Driver}\par{}\citep{DME-Driver} & \COneTable & \DTwoTable\par{}LLaVA & \DOneTable\par{}Custom & \NATable & \COneTable & \DTwoTable\par{}nuScenes & \NATable & \NATable & \COneTable & \DOneTable \\ 
& 29 & \textit{Senna-E2E}\par{}\citep{Senna} & \CThreeTable\par{}Apache2.0 & \DTwoTable\par{}Vicuna v1.5 & \DOneTable\par{}DriveX & \NATable & \CThreeTable\par{}Apache2.0 & \DTwoTable\par{}nuScenes & \NATable & \NATable & \CThreeTable\par{}Apache2.0 & \DTwoTable** \\
& 30 & \textit{DiffVLA}\par{}\citep{diffvla} & \COneTable & \DTwoTable\par{}LLaVA v1.5 & \DTwoTable\par{}NAVSIM v2 & \NATable & \COneTable & \DTwoTable\par{}NAVSIM v2 & \NATable & \NATable & \COneTable & \DOneTable \\
& 31 & \textit{DriveVLM-Dual}\par{}\citep{drivevlm} & \COneTable & \DThreeTable\par{}Qwen-VL & \DOneTable\par{}Custom & \NATable & \COneTable & \DTwoTable\par{}nuScenes & \NATable & \NATable & \COneTable & \DOneTable \\
& 32 & \textit{Solve-E2E-Async}\par{}\citep{solve} & \COneTable & \DTwoTable\par{}LLaVA v1.5 & \DTwoTable\par{}nuScenes & \NATable & \COneTable & \DTwoTable\par{}nuScenes & \NATable & \NATable & \COneTable & \DOneTable \\
& 33 & \textit{DiMA-Dual}\par{}\citep{dima} & \COneTable & \DTwoTable\par{}LLaVA v1.5 & \DTwoTable\par{}nuScenes & \NATable & \COneTable & \DTwoTable\par{}nuScenes & \NATable & \NATable & \COneTable & \DOneTable \\
& 34 & \textit{HE-Drive}\par{}\citep{hedrive} & \COneTable & \DTwoTable\par{}LLaMA 3 & \DTwoTable\par{}nuScenes & \NATable & \COneTable & \DTwoTable\par{}nuScenes & \NATable & \NATable & \COneTable & \DOneTable \\
& 35 & \textit{VDT-Auto}\par{}\citep{vdtauto} & \COneTable & \DThreeTable\par{}Qwen-VL & \DTwoTable\par{}nuScenes & \NATable & \COneTable  & \DTwoTable\par{}nuScenes & \NATable  & \NATable & \COneTable  & \DOneTable  \\
& 36 & \textit{FASIONAD++}\par{}\citep{fasionad} & \COneTable & \UnavailableTable & \DTwoTable\par{}nuScenes & \NATable & \COneTable & \DThreeTable\par{}Bench2Drive & \NATable & \NATable & \COneTable & \DOneTable \\
& 37 & \textit{AsyncDriver}\par{}\citep{chen2024asynchronous} &  \CThreeTable \par{}Apache2.0 &\DTwoTable\par{}LLaMA 2 & \DTwoTable\par{}nuPlan & \NATable & \CThreeTable\par{}Apache2.0  & \DTwoTable\par{}nuPlan & \NATable & \NATable & \CThreeTable\par{}Apache2.0 & \DThreeTable**\\
\bottomrule
\end{tabular}
}
\end{table*}

Open models and datasets play a vital role in the advancement of research and acceleration of practical deployment. They enable researchers to quickly build upon existing work to improve the state-of-the-art, while allowing practitioners to build  high-performing real-world systems without the extensive time needed to reproduce the existing work. It also fosters user trust as transparency reveals strengths and weaknesses of approaches. Moreover, open-source resources lower economic barriers, promoting adaptation in low-income regions. Considering the importance of openness, we present the level of openness for all the 37 approaches for trajectory planning considered in this work.

Following \cite{positionopenness}, we consider training, evaluation and deployment as different stages of building a model, and accordingly score the assets, which is ``data'' and ``code''. For an asset, a higher score implies a higher openness level. Since we study trajectory planning methods that leverage \glspl{FM}, the underlying assets for our use cases require additional elements, as shown in \cref{fig:openness_fig}(a). Specifically, we include\textit{ pretrained \gls{FM} weights} in the training stage as an additional data asset. We also consider driving and language/action datasets separately, as language annotations are generally derived from existing driving datasets, and multiple language annotations can exist for the same driving dataset, e.g., nuCaption~\citep{nucaption}, nuX~\citep{nux}, nuScenes-QA~\citep{NuScenesQA} and DriveLM-nuScenes~\citep{drivelm}.
For the levels of openness, we score  each asset between 1 and 3 to provide targeted information for research and commercial purposes (refer to \cref{fig:openness_fig}(b)), defined as: 
\begin{compactitem}
    \item 1 implies an asset, i.e., code or data, is not publicly released or is released with a license that restricts both non-commercial and commercial usage, including research.
    \item 2 represents that the asset is publicly available and can be used for non-commercial purposes, including research, but not for commercial use. One example license for this is Creative Commons Attribution-NonCommercial 4.0 International.
    \item 3 is for assets that are publicly available and can be used for both research and commercial purposes, e.g., those with Apache License v2.0 or MIT License.
\end{compactitem}

In \cref{tab:openness}, we classify the code and data of all the approaches using these levels of openness. We observe that \textit{there is no approach with all assets being available for both research and commercial purposes.} Only five of the approaches (i.e., Omni-Q, Orion, LMDrive, SimLingo and AgentDriver) released all of their assets openly with some assets limited for commercial usage. Four of these approaches are \glspl{FM} tailored for trajectory planning, and one of them is an \gls{FM} guiding trajectory planning of a modular \gls{AD} model. Consequently, \textit{no open-source implementation is available} for \gls{FM} guiding trajectory planning for an \gls{E2E}-\gls{AD} model (rows 23-36).

\begin{figure}[t]
        \raggedright
        \includegraphics[width=0.95\textwidth]{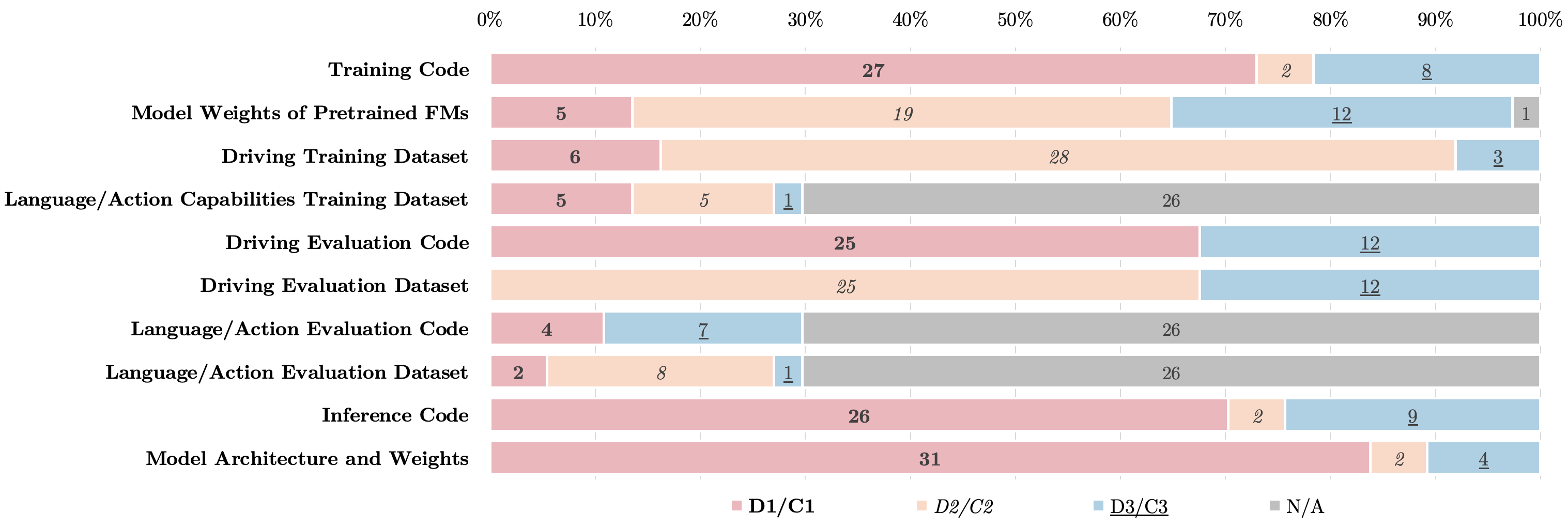}
        \caption{Distribution of openness levels across data and code assets for the 37 surveyed methods. Numerical labels on the bars represent the count of approaches in each openness category per asset.}
        \label{fig:openness_stats}
\end{figure}

To demonstrate and summarise per-asset openness status, we provide the distribution of the openness levels of each asset in \cref{fig:openness_stats}. The figure indicates that the model weights of pretrained \glspl{FM} are commonly open for research (31 out of 37 models), enabling the models to be developed. On the other hand, this doesn't necessarily translate into their derivatives being open. Specifically, \textit{the model architecture and weights is the most restricted asset with only 6 models providing them openly for research and commercial usage (D2 or D3)}. Similarly, as another crucial aspect of reproducibility, \textit{only 10 models openly share the training code (C2 or C3).} This discrepancy creates a significant \textit{bottleneck}: without model parameters and implementation details of the training pipelines, it is difficult for the community to build upon the previous research. Ultimately, this presents a challenge for reproduction and reuse.

We also observe a clear distinction between the openness of synthetic and real-world 
datasets. Synthetic datasets, generated typically via the 
CARLA simulator~\citep{carla}, are often fully-open (D3), allowing for unrestricted use in both research and commercial projects. 
In contrast, the real-world datasets used by the surveyed methods, including nuScenes~\cite{nuscenes} 
and BDD-X~\cite{bddx}, are commonly restricted to non-commercial use cases only (D2). This gap creates friction when translating 
research into industrial practice and may introduce inconsistencies in standardised evaluation across academic and commercial domains.
\section{Open Issues and Emerging Directions}
\label{sec:OpenProblems}
In this section, we identify and discuss the open issues within our scope, while also considering emerging directions in the broader research landscape.

\textbf{Deploying these models can be challenging due to the high inference cost, especially when \gls{CoT} reasoning is used.} Except for a few models using knowledge distillation only during training (\cref{subsec:onlytraining}), all other approaches in our scope require an \gls{FM} for inference. Consequently, mainly due to the large number of parameters and \gls{AR} generation of the output, their latency is commonly longer than what is suitable for \gls{AD}. This is usually accepted as 10-30fps to provide effective responsiveness to the changing environment and to match the latency of sensor data (camera, lidar)~\citep{nuscenes,waymo,womd}.
For example, Orion~\citep{ORION} has more than 7B parameters (see \cref{tab:trajectoryplanningmethods}) and an inference rate of 0.8fps on an Nvidia A800 GPU, even without \gls{CoT}~\citep{oriongithubissuefps}.
Additionally, referring to our previous experiment in \cref{subsubsec:textoutputcot}, the throughput of SimLingo~\citep{SimLingo}, a relatively smaller model than its counterparts, using only the front view, without and with \gls{CoT} reasoning, is 3.6fps and 0.8fps, respectively, on an Nvidia A4500 GPU respectively.
While, efficient processing has been recently considered in a number of models, such as the use of separate slow, reasoning-based FM systems alongside a fast decision system for rapid response times~\citep{fasionad,autovla,wang2025alpamayo}, fast control cycles have generally not been a priority for current models.
Therefore, more efficient approaches are necessary to be able deploy such models in the real-world.

\begin{figure*}[htpb]
        \centering
        \includegraphics[width=0.95\textwidth]{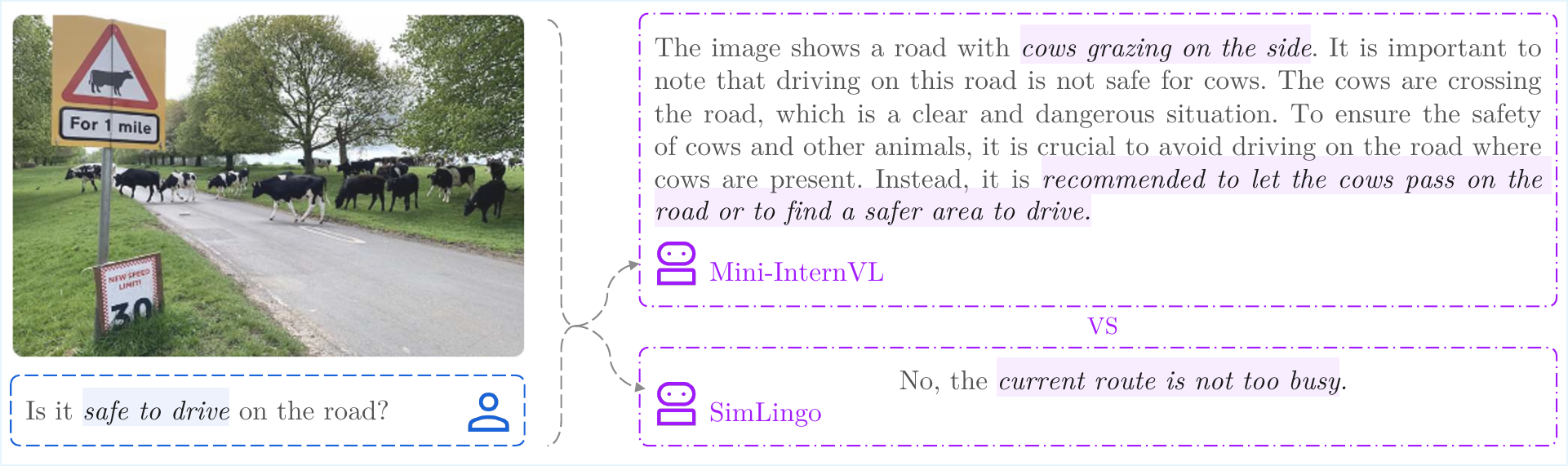}
        \caption{An example showing \textit{vision-language collapse} in the \gls{VLM} of SimLingo after fine-tuning. We first prompt Mini-InternVL, the \gls{VLM} that SimLingo is initialised from, and then SimLingo, i.e., Mini-InternVL after fine-tuning, with the image and text shown in the figure.
        }
        \label{fig:losingcapabilities}
\end{figure*}

\textbf{Fine-tuned \glspl{VLM} become \textit{less capable} of interpreting the world (vision-language collapse).} The superiority of \glspl{FM} tailored for trajectory planning in benchmarks demonstrates that they effectively leverage transfer learning via fine-tuning the pretrained models. However, this fine-tuning procedure can result in the loss of capabilities of the \gls{VLM} that could potentially be more helpful for trajectory planning. This crippling effect might be emerging due to the negative effect of fine-tuning, studied as \textit{concept forgetting}~\citep{mukhotiConceptForgetting} which is an extension of the traditional catastrophic forgetting~\citep{catastrophicforgetting} for fine-tuned \glspl{FM}. We show this effect in \cref{fig:losingcapabilities}, where we prompt the \gls{VLM} used in SimLingo with the same image and text inputs before and after fine-tuning. For comparison purposes, we use one of the inputs we employed for prompting GPT-4o in \cref{fig:motivation}, where GPT-4o provides a comprehensive accurate response. Although it is not as comprehensive as GPT-4o, the response of Mini-InternVL, a relatively small \gls{VLM} that SimLingo is initialised from, is also quite accurate. Specifically, it captures that \textit{the cows are crossing the road} and suggests \textit{either letting the cows pass or finding a safer area to drive}. On the other hand, after fine-tuning, SimLingo provides a \textit{self-contradictory response}, mentioning that it is not safe to drive followed by an explanation that the route is not busy, implying a safe route. Furthermore, SimLingo does not provide any insight about the environment, such as the presence of the cows. Further research and thorough analysis is essential to ensure preservation of necessary capabilities of \glspl{VLM} after fine-tuning on \gls{AD}-specific datasets.

\begin{figure*}[t]
        \centering
        \includegraphics[width=0.95\textwidth]{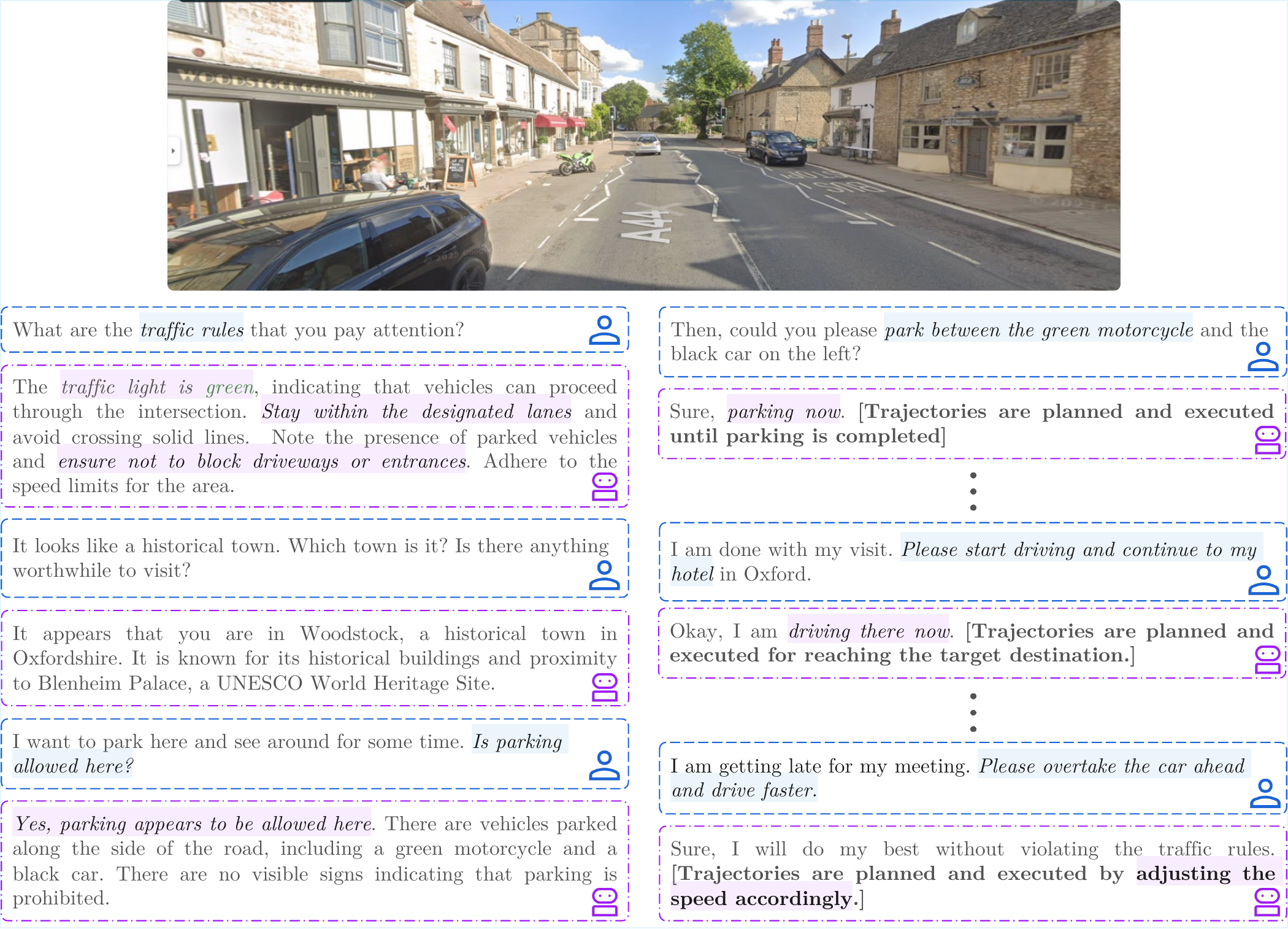}
        \caption{Example \glspl{VQA} and user instructions that the current models lack.
        }
        \label{fig:openissues_capabilities}
\end{figure*}

\textbf{Towards Agentic VLAs.} Existing approaches with language interaction capability generally focus on interpreting the current scene to provide explainability to the user. 
There is however potential to offer further use cases if the question-answer pairs are tailored accordingly. 
For example, the user can ask non-driving-related questions such as the historical details of a landmark in a town that the ego driver passes through, as illustrated in \cref{fig:openissues_capabilities}.
Alternatively, the model could help people with visual impairments understand their surroundings, such as why the traffic is stopping.
One limitation of current models with action interaction capability, is that they are limited to instructions that can be executed in a short time horizon.
DriveMLM and SimLingo consider instructions to change speed or lane, which can be completed within the horizon of a single trajectory output. 
\textit{As a result, the model does not need to remember the instruction over multiple calls, or check if it is completed, which is impractical. 
}%
Though the instruction set used in LMDrive is larger with 56 different types of instructions, it still doesn't require the model to break down a human-like complex instruction into steps as in the example \textit{``please start driving and continue to my hotel in Oxford''}, illustrated in \cref{fig:openissues_capabilities}. 
More sophisticated planning models in the robotics literature~\citep{LLMasZSplanner,smalllanguagemodelsfuture} can help leverage similar capabilities in real-world for \gls{AD} models.

\textbf{Challenges of ensuring safety of action interaction capability in a real-world setup.} Providing action-interaction capability allows substantial flexibility of commands that can be provided by a user. However, due to the open domain of user instructions, there are substantial challenges in ensuring that the vehicle operates safety, regardless of the commands given.  Existing methods do include training of action safety in the dataset, as described in Section~\ref{subsec:fmtailored}, however ensuring safety with an unrestricted command domain remains an open issue. Furthermore, all models investigated in this work with action interaction capability are trained and tested using synthetic setups using CARLA simulator. As domain shift results in a significant drop in performance~\citep{closerlookatdomainshift,domaingeneralizationsurvey,saod,kuzucu2024calibration}, deploying existing models without addressing sim-to-real gap is impractical. Therefore, designing methods to address sim-to-real gap and curating real-world training datasets is necessary to be able develop, and eventually deploy, models with action interaction capability. 

\textbf{Further analyses are necessary to identify what is significant for improving driving performance.} There are major differences in model and experimental design choices across methods, which makes it difficult to understand why a method makes a difference. For example, Orion is trained on the B2D-Base dataset~\citep{bench2drive} with around 7 hours of video clips, combines an off-the-shelf vision encoder with a custom Q-Former with memory blocks for videos, trains the model in three stages, has more than 7B parameters, uses a variational autoencoder as trajectory planning head, and employs a comprehensive \gls{CoT} including scene description, scene analysis, action reasoning, and history review. This models achieves $77.74$ driving score on the closed-loop B2D benchmark, significantly outperforming \gls{E2E}-\gls{AD} approaches such as VAD with $42.35$ driving score~\citep{bench2drive}. On the other hand, SimLingo employs Mini-InternVL (1B parameters) as an off-the-shelf \gls{VLM}, fine-tunes it on more than 200 hours of video in a single stage relying on a data resampling approach coined as bucketing, uses a different set of language annotations and \gls{CoT}, and finally achieves a driving score exceeding $85$. These major differences between models, training approaches as well as training datasets make it difficult to understand why a model makes a difference. As a result, a deeper understanding of how these factors affect the model is necessary. 

\textbf{Lack of standardised benchmarks and metrics for FM-based trajectory planning.} Building upon the aforementioned challenges, the establishment of standardised benchmarks and metrics is essential to facilitate rigorous and equitable comparison across different approaches, and to systematically identify the key factors contributing to performance improvements (e.g., training data characteristics, architectural design choices, model scale). Such benchmarks and evaluation metrics should address multiple dimensions of model performance, including:
\begin{compactitem}
    \item \textit{Driving performance under resource-constrained settings.} \gls{FM}-based approaches for \gls{AD} exhibit considerable variation in model size. For instance, SimLingo comprises 1B parameters~\citep{SimLingo}, while Orion contains approximately 7B parameters~\citep{ORION}. Consequently, inference latency varies substantially across methods, further influenced by design decisions such as \gls{CoT} usage. Therefore, evaluating driving performance under controlled resource constraints (e.g., bounded inference time) becomes crucial for assessing the practical viability of \gls{FM}-based methods in real-world deployment scenarios.
    \item \textit{Language interaction and action interaction capabilities.} Beyond core driving competencies, certain \glspl{FM} designed for trajectory planning incorporate additional capabilities for language and action interactions. To comprehensively evaluate these capabilities, standardised benchmarks encompassing both real-world and simulated data are necessary.
    \item \textit{Reasoning of the models.} Furthermore, analogous to the common benchmarks used for evaluating \glspl{FM}, standard benchmarks for evaluating the reasoning capabilities of \glspl{FM} tailored for trajectory planning are needed. Drawing from the broader \gls{FM} literature, the ARC-AGI benchmark~\citep{arcprize2024technical} exemplifies this approach by designing visual reasoning tasks that humans solve intuitively while \glspl{FM} struggle. Similar domain-specific benchmarks should be developed for \gls{AD} to assess reasoning abilities when processing complex linguistic inputs, as well as multimodal language-vision grounding capabilities. Developing such benchmarks presents substantial challenges and warrants dedicated attention from the research community.
\end{compactitem}

\textbf{Improving driving and evaluation with World Models.} Training current \gls{FM}-based methods for \gls{AD} typically relies on recorded driving demonstrations from an expert, which can be either human-generated or derived from another model with access to privileged information. Consequently, model performance is fundamentally bounded by the capabilities of the expert, and the availability of training data remains inherently limited. Furthermore, existing closed-loop simulators exhibit significant constraints: they are either characterised by low fidelity, as exemplified by CARLA~\citep{carla}, or impose substantial computational demands, as in the case of novel viewpoint synthesis approaches~\citep{kerbl20233d}.  In contrast, world models aim to \textit{learn} the underlying dynamics of the autonomous vehicle's operating environment, thereby enabling the generation of plausible future states. Therefore, world models can be leveraged to synthesise novel training data~\citep{ren2025cosmos, Zhao_2025_CVPR} as well as to facilitate evaluation procedures that emphasise rare and safety-critical edge cases, thus addressing long-tail distributional challenges. Additionally, world models can also be integrated with driving models, allowing world dynamics to influence driving decisions~\citep{wang2024driving}, for example by rolling out possible futures using the world model and selecting the most suitable trajectory for driving.
\section{Conclusive Remarks}
\label{sec:Conclusions}
In this paper, we provided a comprehensive review of trajectory planning methods that utilise an \gls{FM}. To offer a complete and coherent perspective, we introduced a taxonomy of these methods based on how an \gls{FM} is employed. Using this taxonomy, we discussed the corresponding approaches separately in a detailed and comparative manner, providing a unified yet critical point of view. We also investigated the openness of the approaches to assist practitioners and researchers in selecting suitable models. Finally, we identified open issues that are critical for developing practical models with the desired functionalities. With this review, the community can better understand the current state of the field and the directions to pursue for developing more capable solutions for \gls{AD} leveraging \glspl{FM}.

\blockcomment{
\subsubsection*{Broader Impact Statement}
In this optional section, TMLR encourages authors to discuss possible repercussions of their work,
notably any potential negative impact that a user of this research should be aware of. 
Authors should consult the TMLR Ethics Guidelines available on the TMLR website
for guidance on how to approach this subject.

\subsubsection*{Author Contributions}Observation
If you'd like to, you may include a section for author contributions as is done
in many journals. This is optional and at the discretion of the authors. Only add
this information once your submission is accepted and deanonymized. 

\subsubsection*{Acknowledgments}
Use unnumbered third level headings for the acknowledgments. All
acknowledgments, including those to funding agencies, go at the end of the paper.
Only add this information once your submission is accepted and deanonymized. 
}
\bibliography{references}
\bibliographystyle{tmlr}

\appendix
\clearpage

\section*{APPENDIX}
\renewcommand{\thefigure}{A.\arabic{figure}}
\renewcommand{\thetable}{A.\arabic{table}}
\renewcommand{\theequation}{A.\arabic{equation}}
\renewcommand{\thesection}{A}
\section{Further Details on the Openness of the Methods}
\label{app:conventions}

We classify and discuss the openness of the assets of each method in \cref{sec:openness}. Here, we include the conventions, which also clarify few outlier cases in the openness classification in \cref{tab:openness}:
\begin{itemize}
    \item If evaluation of an approach is performed on multiple datasets, the dataset with \textit{the least restrictive license} is considered for the classifying ``Driving Evaluation Dataset'' and ``Language/Action Evaluation Dataset'' assets. The name of the dataset in question is noted as an explanation. This selection is chosen to reflect when at least part of the results can be reproduced by the community.
    \item On the other hand, if training is performed on multiple datasets jointly, the dataset with \textit{the most restrictive license} is considered for classifying ``Trajectory Planning Training Dataset'' and ``Language/Action Capabilities Training Dataset` assets. The name of the dataset in question is noted as an explanation. This convention is chosen as all datasets within a mixture are needed to reproduce the results. However, few methods, e.g. Drive-VLM~\citep{drivevlm}, train the models on different dataset mixtures, with results reported independently. In such a case, the score of \textit{the least restrictive} mixture of datasets is used for the corresponding cell entry. 
    \item To provide further clarity on the basis of our classification, we include the license name for the code assets, and the dataset name for the data assets (except model weights).
    \item For Omni-Q~\citep{omnidrive}, Orion~\citep{ORION} and AsyncDriver~\citep{chen2024asynchronous}, the license of the released model weights is in conflict with pretrained \glspl{FM} weights. We classify these entries based on the released model license and warn the reader that additional restrictions may apply (noted as *).
    \item For LMDrive~\citep{lmdrive} and Senna-E2E~\citep{Senna}, there are two contradicting licenses for the model weights, in the same huggingface space. We classify the entry based on the more restrictive license which also aligns with the openness score of the pretrained \glspl{FM} weights (noted as **).
    \item If no license is specified, default copyright law applies, which is highly restrictive.
    \item If the model combines a vision encoder and an LLM--instead of using an off-the-shelf VLM--we consider the most restrictive license for the pretrained \glspl{FM} weights classification. Note, any components trained from scratch are ignored (e.g. \gls{LLM} in CarLLaVA~\cite{CarLLaVA})
    \item Closed-loop evaluation benchmarks like CARLA Town05~\citep{prakash2021multi}, CARLA Longest6 \citep{chitta2022transfuser}, Bench2Drive~\citep{bench2drive} and LAV~\citep{lav} are considered within the "Driving Evaluation Dataset" asset as they provide route splits or auxiliary scenario configuration files.
\end{itemize}

\blockcomment{
\section{VLMs Supporting Autonomous Driving}
\label{sec:supportplanning}
Though we haven't covered in the paper, there are also different aspects that the \glspl{VLM} help \gls{AD} models. For the sake of completeness, we discuss them below (\KO{There should be more tasks and methods here}):

\paragraph{VLMs helping the control task}
\begin{itemize}
    \item LanguageMPC: Large Language Models as Decision Makers for Autonomous Driving \citep{languagempc}
    \item ChatGPT as Your Vehicle Co-Pilot: An Initial Attempt \citep{copilot}
\end{itemize}

\paragraph{AD VLMs that yield a limited set of meta actions}
\begin{itemize}
    \item Drive like a human: Rethinking autonomous driving with large language models, WACV 2024 - it doesn't output a trajectory or control signals, just a limited set of 5 meta actions. So, we can mention but not sure we should include \citep{drivelikeahuman}.
    \item Empowering Autonomous Driving with Large Language Models: A Safety Perspective, Workshop on LLM Agents ICLR2024 - Similar to Drive like a human, meta actions only \citep{safetyagent}
    \item  Dilu: A knowledge-driven approach to autonomous driving with large language models - same line of work, only 5 meta actions \citep{dilu}
    \item AlphaDrive: Unleashing the Power of VLMs in Autonomous Driving via Reinforcement Learning and Reasoning - meta actions only \citep{alphadrive}
    \item LaMPilot: An Open Benchmark Dataset for Autonomous Driving with Language Model Programs, CVPR 2024  \cite{lampilot} - instruction to code, might be included in the taxonomy but not sure
    \item SurrealDriver: Designing LLM-powered Generative Driver Agent Framework based on Human Drivers’ Driving-thinking Data, IROS 2024 \cite{SurrealDriver}
    \item Embodied Understanding of Driving Scenarios \cite{embodiedunderstanding}
    \item RAD: Retrieval-Augmented Decision-Making of Meta-Actions with Vision-Language Models in Autonomous Driving \citep{RAD}
\end{itemize}

\paragraph{AD VLMs for VQA}
\begin{itemize}
    \item Drive as You Speak: Enabling Human-Like Interaction with Large Language Models in Autonomous Vehicles, WACV 2024 - interesting paper, only outputs textual data, not clear how to convert to control signals \citep{days}
    \item Can LVLMs Obtain a Driver’s License? A Benchmark Towards Reliable AGI for Autonomous Driving, AAAI 2025 - Just finetune and prompt the model to generate the waypoints. They use Qwen-VL- chat. Not sure if we should include. We can mention it in DriveVLM part as they follow that paper to show the benefit of their dataset \cite{VLMdrivinglicense}
    \item Holistic Autonomous Driving Understanding by Bird's-Eye-View Injected Multi-Modal Large Models, CVPR 2024 \cite{HolisticUnderstanding}
    \item Reason2Drive: Towards Interpretable and Chain-based Reasoning for Autonomous Driving. ECCV 2024 \cite{reason2drive}
    \item Dolphins: Multimodal Language Model for Driving, ECCV 2024 \cite{dolphins}
    \item Human-Centric Autonomous Systems With LLMs for User Command Reasoning, WACV 2024 Workshops \cite{humancentricautonomoussystemsllms}
\end{itemize}
}

\end{document}